\documentclass{article}

\PassOptionsToPackage{numbers, compress}{natbib}

\PassOptionsToPackage{numbers, sort}{natbib}
\usepackage[final]{neurips_2022}

\usepackage{url}            %
\usepackage{xcolor}         %
\usepackage{pifont}%
\usepackage{comment}
\usepackage{amsmath}
\usepackage{booktabs}
\usepackage[pagebackref,breaklinks,colorlinks]{hyperref}      %

\usepackage{booktabs}       %
\usepackage{amsfonts}       %
\usepackage{nicefrac}       %
\usepackage{graphicx}
\usepackage{microtype}      %
\usepackage[T1]{fontenc}    %
\usepackage[utf8]{inputenc} %
\newcommand{\xmark}{\ding{55}}%
\usepackage{minitoc}

\newcommand{\revised}{\textcolor{black}}
\newcommand{\camera}{\textcolor{black}}
\title{Segmenting Moving Objects via an \\ Object-Centric Layered  Representation}

\author{%
  Junyu~Xie$^1$\hspace{0.8cm} Weidi Xie$^{1,2}$ \hspace{0.6cm} Andrew Zisserman$^{1}$ \\
  $^1$Visual Geometry Group, Department of Engineering Science, University of Oxford, UK  \\
  $^2$Coop. Medianet Innovation Center, Shanghai Jiao Tong University, China\\
  \small \texttt{\{jyx,weidi,az\}@robots.ox.ac.uk} \\
   \small \href{https://www.robots.ox.ac.uk/~vgg/research/oclr/}{\texttt{https://www.robots.ox.ac.uk/\textasciitilde vgg/research/oclr/}}\\
}

\begin{document}
\maketitle

\begin{abstract}
\label{sec:abstract}
The objective of this paper is a model that is able to discover, track and segment multiple moving objects in a video. We make four contributions:
First, we introduce an object-centric segmentation model with a depth-ordered layer representation. 
This is implemented using a variant of the transformer architecture that ingests optical flow, where each query vector specifies an object and its layer for the entire video. The model can effectively discover multiple moving objects and handle mutual occlusions;
\revised{Second, we introduce a scalable pipeline for generating multi-object synthetic training data via layer compositions, that is used to train the proposed model, 
significantly reducing the requirements for labour-intensive annotations, and supporting Sim2Real generalisation;}
Third, we conduct thorough ablation studies, 
showing that the model is able to learn object permanence and temporal shape consistency, and is able to predict amodal segmentation masks; 
\revised{Fourth, we evaluate our  model, trained  only on synthetic data, on standard video segmentation benchmarks, DAVIS, MoCA, SegTrack, FBMS-59, 
and achieve state-of-the-art performance among existing methods that do not rely on any manual annotations.}
With test-time adaptation, we observe further performance boosts. 
\end{abstract}

\section{Introduction}
\label{sec:intro}
\vspace{-0.1cm}
Humans have the ability to discover and segment moving objects in videos. This ability is present even in young infants, 
together with notions of object permanence through occlusions, 
and temporal shape constancy~\cite{Piaget54,Baillargeon85,Spelke90,Baillargeon91}. 
Achieving this ability by machine has long been a goal of the field~\cite{Brox10,Ochs11,Keuper15,Bideau16,Tokmakov17, Bideau18,xie19, Tokmakov19, Dave19, yang_loquercio_2019,Lamdouar20,Yang21a}, 
and modern methods are now
able to reliably discover and segment single moving objects using both appearance (RGB) and motion (optical flow) streams of the video. 
However, success is limited in the case of multiple moving objects and occlusion.

\begin{figure}[t]
  \hspace{-0.6cm}
  \includegraphics[width=14.5cm]{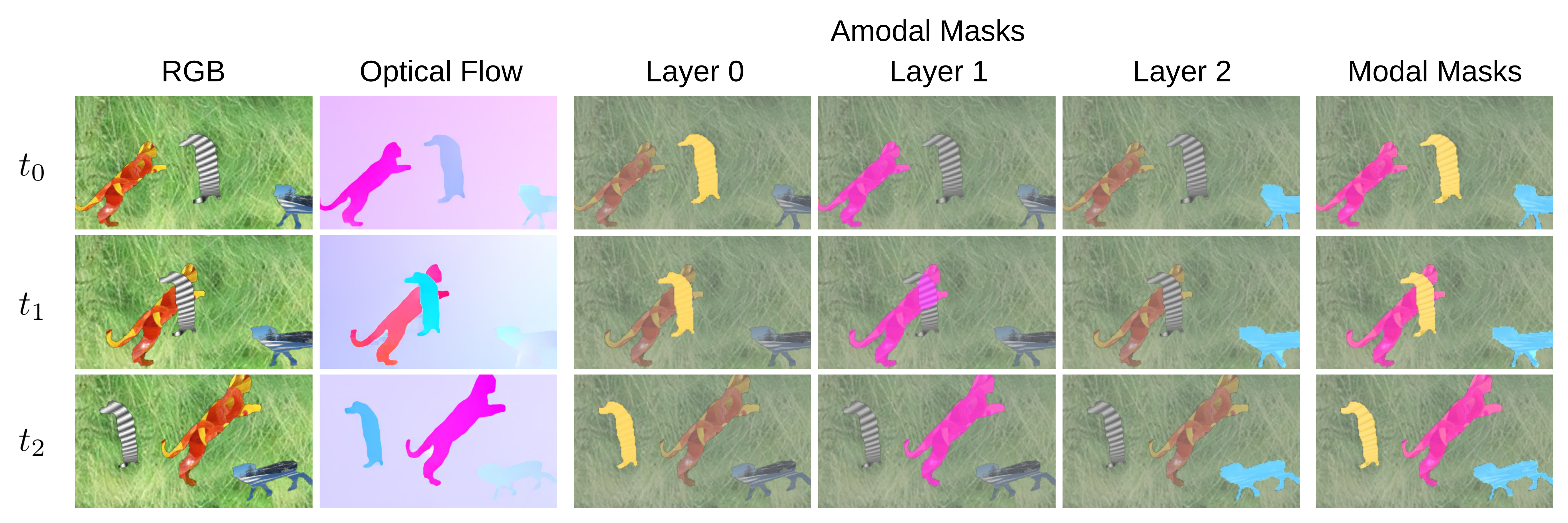}
  \vspace{-0.6cm}
  \caption{\textbf{Multiple-object Discovery and Segmentation.} We develop a layered model for segmenting and tracking multiple objects by their motion, even under mutual occlusion. 
  The model takes only optical flow as its input, 
  and predicts an \textbf{amodal} segmentation~(whole physical structure) for each object, and associating an object to the same layer throughout the video. 
  The resultant \textbf{modal} masks~(visible parts) can be constructed by overlaying layers based on the inferred depth ordering~(in this case, layer 0-1-2 from front to back). From left to right: RGB sequence, estimated (imperfect) optical flows, amodal mask inferences from our model and resultant modal masks.}
  \label{fig:teaser}
  \vspace{-.4cm}
\end{figure}

Our goal in this paper is to develop and study a model that can discover multiple moving objects and also handle their mutual occlusions. 
To this end, we propose a {\em layered} object-centric model that predicts {\em amodal} segmentation masks~\cite{zhu2017semantic} for each object throughout the video.
Note that,  given amodal segmentations, 
the depth ordering of the object layers can be determined from occlusions in the observed (modal) frames when the objects overlap.
See Figure~\ref{fig:teaser}.
Our innovation is to build this mechanism into the representation: 
when objects overlap in a frame, 
the model has to maintain the amodal shapes through temporal consistency,
and can thus infer depth ordering.

To achieve this goal we use only optical flow as our primary information stream. There are benefits and drawbacks to this choice. 
The benefits are two fold: 
first, the information that we are after, the segmentation
masks, are directly available from discontinuities in the flow field; second, a key advantage of using solely optical flow is that the domain
adaptation problem is largely avoided~\cite{doersch2019sim2real, Lamdouar21} -- and this means that the model can be trained on synthetic sequences and applied directly to real sequences without a significant Sim2Real disparity.
However, the drawbacks are also two fold: first, unlike the appearance stream, objects can be invisible in flow when they are not moving, 
thereby requiring the model to build a concept of object permanence;
and second, but more subtly, flow is not like appearance in that we cannot expect constancy even when the object is moving~(for example, due to 
camera shake that affects the flow but not the appearance), and thus the model cannot strongly use flow coherence as a cue but must instead learn to model the segmentation mask shape.

We make four contributions: (i) we introduce an
 \textbf{O}bject-\textbf{C}entric  \textbf{L}ayered \textbf{R}epresentation (\textbf{OCLR}) model
for discovering and segmenting multiple moving objects and inferring their mutual occlusions from optical flow alone.
The architecture of the model is based on a transformer, 
with DETR like learnable queries, where each query determines the amodal segmentation mask for an object and its layer order throughout the video.
\revised{(ii) we introduce a layer-based synthetic pipeline to generate multi-object training sequences that cover the variability of the target real sequences in terms of: multiple deforming objects, overlapping objects in some frames, moving cameras, and non-static backgrounds. }
(iii) We demonstrate that the model trained on flow from the synthetic videos is able to learn object permanence and temporal shape constancy --
even if an object is partially occluded in many frames, the model is still able to predict an amodal mask.
\revised{(iv) Finally, we evaluate the model~(trained only on synthetic sequences) on multiple real video segmentation benchmarks, DAVIS, MoCA, SegTrack, and FBMS-59; 
and demonstrate strong Sim2Real generalisation, achieving superior segmentation performance compared to other human-label-free methods.}
The addition of a test-time adaptation to include appearance from
a self-supervised DINO model boosts the performance further, even outperforming some  supervised approaches that are finetuned on real data sequences.
\section{The Object-Centric Layered Representation (OCLR) Model}
\label{sec:model}
\vspace{-0.1cm}

This section describes our object-centric layered representation model that ingests
optical flow for the frames of a video, and outputs the segmentation masks and tracks of the moving objects it contains. The model also determines the number of moving objects and their depth order. Specifically, the model predicts amodal ({\em i.e.}\ complete) segmentation masks and associates each object with a layer. 
The layers are composed to form the modal segmentations in the observed optical flow. The model is trained on synthetic sequences to learn temporal coherence and maintain the object shape even under occlusion, and as a consequence, develops a notion of object permanence.

\vspace{-0.05cm}
\subsection{Layered representation}
\label{sec:layer}
\vspace{-0.1cm}
For a given video with $T$ frames, 
we can represent its motion by optical flows, 
$\mathcal{V}_{\text{flow}} = \{F_1, F_2, \dots, F_T\}$,
we aim to segment the moving objects in videos by adopting a layered model, where each object is segmented and consistently associated to one layer throughout the video:
\begin{equation}
\{(\hat{\mathcal{A}}_1, \hat{r}_1), \dots, (\hat{\mathcal{A}}_{T}, \hat{r}_{T})\}= \Phi(\mathcal{V}_{\text{flow}})
\end{equation}
where $\hat{\mathcal{A}}_i \in \mathbb{R}^{H \times W \times K}$ refers to the binary \textbf{amodal} segmentation of the $i$-th frame, together with a predicted layer depth $\hat{r}_i \in \mathbb{R}^K$.
$H, W$ refer to the input height and width, and $K$ denotes the number of object layers. Note that, the layer depth value here does not have a true meaning, but only indicates the layer ordering.
By definition, at all time steps, 
the same layer would remain empty or always bind to the \textbf{same} object throughout the video, {\em i.e.}~objects can be effectively tracked by the considered layered representation.

{\noindent \bf Modal mask generation. }
Given the predicted amodal segmentation and layer depth, 
we describe the composition procedure to generate the \textbf{modal} segmentation for moving objects.
In specific, we permute the index of layers~(\textbf{amodal} segmentations) based on their depth values, such that, $\hat{\mathcal{A}}^{k-1}$ is always in front of $\hat{\mathcal{A}}^k$. \revised{Note that, pixels in amodal masks defined here are binary-valued, {\em i.e.}~either 0 or 1.} For the $i$-th frame, 
its modal segmentation can thus be computed via an iterative, front-to-back blending procedure :
\begin{equation}
\hat{\mathcal{M}}_i^k = (1 - \alpha_i^k) \odot \hat{\mathcal{A}}_i^k 
\end{equation}
where \revised{$\alpha_i^k = \text{clip}(\alpha_i^{k-1} + \hat{\mathcal{M}}_i^{k-1}, 0, 1)$},
denoting an accumulated opacity layer with \revised{$\alpha_i^{0} = 0$}.
\vspace{-0.05cm}
\subsection{Architecture}
\label{sec:arch}
\begin{figure}[t]
  \includegraphics[width=\textwidth]{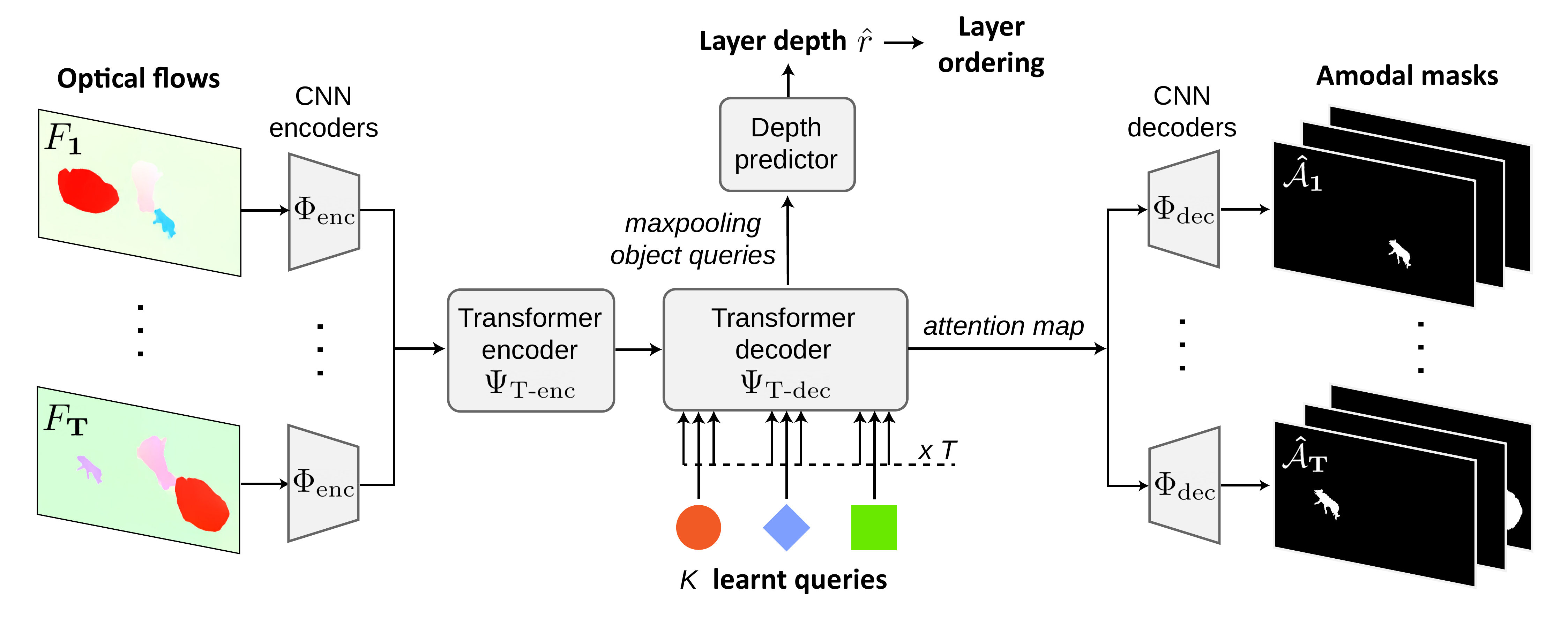}
  \caption{\textbf{OCLR Model architecture.} 
  The model is a U-Net architecture with a Transformer bottleneck.
  It takes optical flow as input and extracts spatial features by CNN encoders. A transformer encoder jointly processes spatio-temporal features across all frames, followed by a transformer decoder that determines the layer representations. Each learnable query vector in the  transformer decoder is associated to one object and used to infer its layer depth.
  Additionally, the cross-attention maps from the last transformer decoder are extracted and upsampled by CNN decoders to infer amodal segmentation for the moving objects.
  Note, the skip connections from the CNN encoders to decoders are not shown.}
  \label{fig:arch}
  \vspace{-0.5cm}
\end{figure}

\vspace{-0.1cm}
In order to accommodate the layered representation, 
we introduce an architecture that takes optical flows as input,
and outputs a set of layers and their depths, see Figure~\ref{fig:arch}.
We adopt a U-Net architecture with a Transformer-based bottleneck, 
the queries in transformer decoder play the role as layers in our proposed representation, each is either empty or encoding the amodal shape of one moving object along the video.

{\noindent \bf CNN backbone. }
Given a sequence of optical flow inputs,
a U-Net encoder is used to compute frame-wise features,
$\mathcal{V} = \{v_1, \dots, v_{T}\} = \{\Phi_{\text{enc}}(F_1), \dots, \Phi_{\text{enc}}(F_{T})\}$,
where $v_i \in \mathbb{R}^{h \times w \times c}$ refers to the feature maps, with height $h$, width $w$ and number of feature channels $c$.
Following~\cite{Yang21a}, we use Instance Normalisation (IN)~\cite{ulyanov2016instance} after each convolution, 
which encourages the separation of foreground motions from background.

{\noindent \bf Transformer encoder. }
We flatten the outputs from the ConvNets in both temporal and spatial dimensions, and inject the positional information by adding a set of learnable spatio-temporal embeddings. 
As a consequence, 
the video features are converted into a sequence of vectors,
and passed into Transformer Encoder to model the temporal relationship across multiple frames.

{\noindent \bf Transformer decoder. } 
We pass the vector sequence to transformer decoder,
where $K$ learnable embeddings are used as object queries for individual layers.
We broadcast the queries along the temporal dimension, 
and supply positional information with learnable temporal embeddings,
$\mathcal{Q}_{\text{obj}} \in \mathbb{R}^{TK \times c}$.
The queries that only differ in temporal embeddings 
will be associated to the same moving object along the video.

As outputs from transformer decoder, 
we extract the last cross-attention maps and recover their spatial dimensions, $\mathcal{\rho} \in \mathbb{R}^{TK \times h \times w \times \text{nheads}}$, which will be progressively upsampled with skip connections from the ConvNets encoders, resulting in binary amodal segmentation for the object queries in each frame,
$\{\hat{\mathcal{A}}_1, \dots, \hat{\mathcal{A}}_{T}\} = \Phi_{\text{dec}}(\mathcal{\rho})$.

{\noindent \bf Layer ordering. } 
Apart from amodal segmentations, the model will also infer the depth of the layers. 
We assume a global layer ordering that remains fixed across frames. As shown in Figure~\ref{fig:arch}, output object queries from transformer decoder are maxpooled along the temporal dimension and passed through two feed-forward layers, to get the output $\hat{r} \in \mathbb{R}^{1 \times K}$, indicating the depth for each of the $K$ layers. Intuitively, this is equivalent to reason the layer depth based on one selected key occlusion frame from the whole sequence. By defining the lowest depth value as corresponding to the top layer, we then obtain a depth order prediction between layers. 
Note that, for layers that are empty or contain non-interacting objects, the ordering is not strictly defined, 
thus any prediction should not be penalised.

\vspace{-0.05cm}
\subsection{Training objectives}
\label{sec:objectives}
\vspace{-0.1cm}
In this section, we describe the training procedure for the proposed architecture, with groundtruth supervision on both amodal segmentation and layer ordering. 

{\noindent \bf Amodal segmentation loss. } 
In addition to the conventional pixelwise binary classification~($\mathcal{L}_{\text{bce}}$) on each of the $K$ layers, we also adopt a pixelwise classification loss defined only on a strip region around object boundaries, termed as the boundary loss $\mathcal{L}_{\text{bound}}$. This additional loss further emphasises the boundary regions, forcing the model to maintain shape information.
During training, since the predicted amodal segmentations are permutation invariant, 
we use Hungarian matching to match their corresponding ground truth~\cite{carion2020end},
and then compute the amodal loss :
\begin{equation}
    \mathcal{L}_{\text{amodal}} = \frac{1}{KT}\sum_{k=1}^K\sum_{i=1}^T\left( \lambda_{\text{bce}}\cdot\mathcal{L}_{\text{bce}}(\hat{\mathcal{A}}^k_i, \mathcal{A}^k_i) +  \lambda_{\text{bound}}\cdot\mathcal{L}_{\text{bound}}(\hat{\mathcal{A}}^k_i, \mathcal{A}^k_i)\right)
\end{equation}
where $\lambda_{\text{bce}}$ and $\lambda_{\text{bound}}$ are loss weights. 

{\noindent \bf Layer ordering loss. } 
For layers with mutual occlusions, we train the order prediction module with ranking losses on all $K!$ pairs of layers, formally,
\begin{equation}
    \mathcal{L}_{\text{order}} = \frac{1}{K!}\sum_{i\not = j}-\log\left(\sigma{\left(\frac{\hat{r}_i - \hat{r}_j}{\tau}\right)}\right)
\end{equation}
where $\sigma$ is the sigmoid function with a temperature factor $\tau$, and the ground-truth ordering indicates a relative depth relationship $\hat{r}_i > \hat{r}_j$ between layer $i$ and $j$.%

{\noindent \bf Total loss. } 
The overall loss for training is a combination of amodal $\mathcal{L}_{\text{amodal}}$ and ordering loss $\mathcal{L}_{\text{order}}$:
\begin{equation}
\mathcal{L}_{\text{total}} = \mathcal{L}_{\text{amodal}} + \lambda_{\text{order}} \cdot \mathcal{L}_{\text{order}}
\end{equation}
where $\lambda_{\text{order}}$ is the weight factor for the layer ordering loss. 

{\noindent \bf Discussion. }
As discussed in early sections, our model is trained to infer amodal ({\em i.e.}~complete) object segmentations and layer depth, even though the object itself may sometimes be invisible in the flow field due to being temporarily static, or partially occluded by other objects.
As a result, the model is forced to always maintain the object shape internally, {\em i.e.}~a notion of object permanence, 
as well as infer mutual occlusions from the visible flows.
In the next section, we will introduce a pipeline for simulating video sequences, where groundtruth amodal segmentations and layer orderings can be generated at scale to train our model.

\section{Synthetic dataset generation}
\label{sec:syn}
\vspace{-0.1cm}
\begin{figure}[t]
  \hspace{-0.2cm}
  \includegraphics[width=\textwidth]{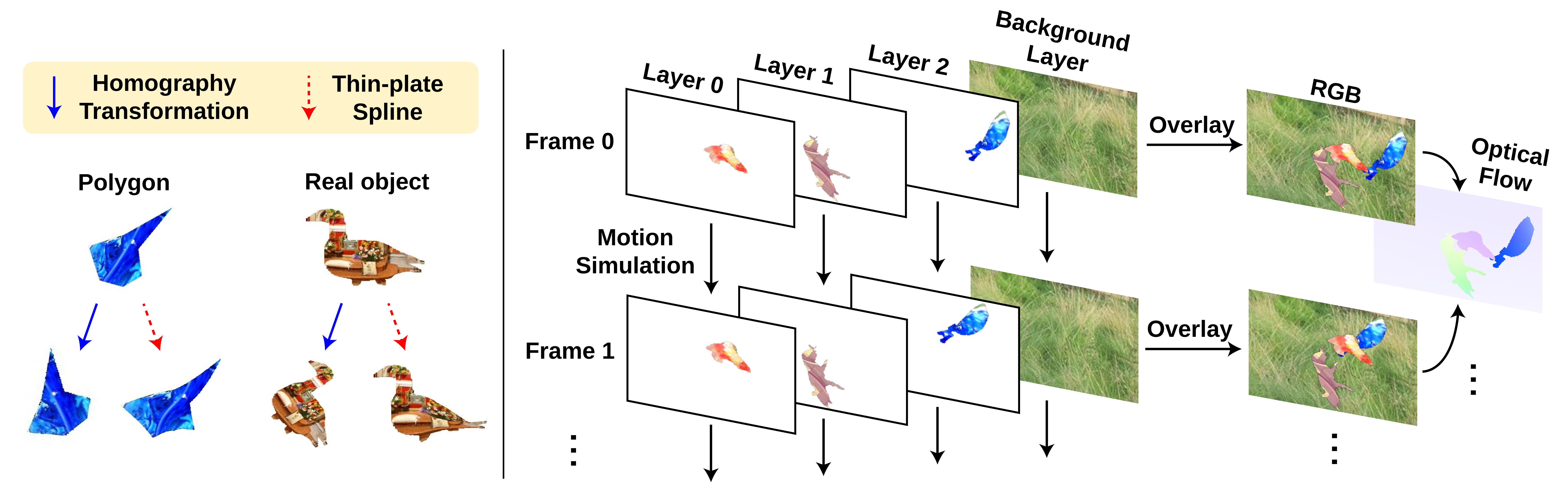}
  \vspace{-0.2cm}
  \caption{\textbf{Overview of the synthetic data pipeline.} Left: Motion simulations of polygon and real object sprites based on homography transformations (including spatial translations) and thin-plate spline.  Right: Layer composition of frames with independently moving objects.}
  \label{fig:pipeline}
  \vspace{-.5cm}
\end{figure}

We introduce a scalable pipeline to synthesize videos to train the proposed layered object-centric model.
The pipeline builds on the method of Lamdouar {\it et al.}~\cite{Lamdouar21}, 
but extends it to simulating videos with multiple objects and complex inter-object occlusions through a layer composition.

As shown in Figure~\ref{fig:pipeline}, 
we first simulate independent motions for foreground objects and backgrounds within each layer. 
To construct a frame, 
we overlay the layers via an iterative binary blending procedure,
and an occlusion occurs when objects in different layers reside in the same spatial position, similar to the idea discussed in Sect.~\ref{sec:layer}.
The proposed pipeline yields rich information at frame level, 
including RGB appearances, groundtruth optical flows, and modal and amodal object segmentations.
Additionally, we can also compute optical flows for the RGB sequences with an off-the-shell flow estimator, for example, RAFT~\cite{Teed20}. 
In the following sections, 
we describe some key procedures for video simulation,
and leave more technical details to the Supplementary Material.

{\noindent \bf Foreground objects. }
\label{sec:pipeline}
We generate opaque 2D sprites with random shapes and RGB textures as foreground objects. 
In detail, we adopt two major shapes, namely polygons, and real object sprites. The polygon shapes can be either convex or non-convex, with different numbers of sides. A minimum distance between vertices is also defined to better support vertex-based deformations. On the other hand, the real object masks are directly sourced from the silhouettes in the YouTubeVOS dataset~\cite{Xu18-YTB}. 
These generated objects are then applied with textures sampled from the PASS dataset~\cite{Asano21}, which is a large-scale image dataset with human identifiable information removed. 

{\noindent \bf Non-rigid object transformations. }
To simulate non-rigid motions,
we apply both homography transformations and thin plate splines to the objects. Thin plate splines define a non-linear coordinate transformation with a set of control points. For polygon-shaped objects, 
we consider their vertices as control points, 
and spatially perturb a subset of the vertices by distances $d_i\in(0, D_i / 2)$, where $D_i = \min{(|\mathbf{P}_i - \mathbf{P}_j|,\forall j)}$, defines the minimum distance between the perturbed point $\mathbf{P}_i$ and any other points $\mathbf{P}_j$. 
For real object silhouettes, 
a uniform grid is distributed over the object, 
and the grid points are defined as control points,
with $D_i$ being the distance between adjacent grid points.

{\noindent \bf Backgrounds and artificial occluders. }
To simulate the camera motions,
we randomly sample images from the PASS dataset, 
and apply random homography transformations and color jitterings.
Additionally, we replace backgrounds with real videos cropped from copyright-free sources to simulate the complex background motions and depth variations. 
For cases with objects being occluded, 
we introduce artificial occluders superimposed as the top layer. These occluders are designed to have the same motion as the background.

{\noindent \bf Summary.}
The proposed simulation pipeline enables an arbitrary number of videos to be generated, together with multiple types of groundtruth annotation; for example, amodal segmentation and layer ordering. We prepare $4$k synthetic sequences (around $120$k frames) for training, the video sequences contain $1$, $2$, or $3$ objects in equal proportions. In the following, we train all models on this synthetic video dataset unless otherwise specified.
\vspace{-0.2cm}

\section{Experiments}
\label{sec:experiment}
\vspace{-0.15cm}
\subsection{Datasets}
\vspace{-0.1cm}
To evaluate our multi-layer model, 
we benchmark on several popular datasets for video object segmentation tasks. A brief overview of the datasets is given below, with full details in the Supplementary Material.

For single object video segmentation, 
we evaluate the model on  DAVIS2016~\cite{Perazzi16}, SegTrackv2~\cite{FliICCV2013}, FBMS-59~\cite{OB14b} and MoCA~\cite{Lamdouar20}. 
Note that, despite the fact that multiple objects may be annotated, the community often treats SegTrackv2 and FBMS-59 as a benchmark for single object segmentation~\cite{Jain17,yang_loquercio_2019} by grouping objects in the foreground.

To benchmark motion-based segmentation for multiple objects, we introduce a synthetic validation dataset (Syn-val) and a curated dataset (DAVIS2017-motion). The former is generated with the same parameters as our synthetic training set~(Sect.~\ref{sec:syn}), containing over $300$ multi-object sequences~(around $10$k frames) with $1,2,3$ objects at equal proportions, and controllable occlusions for evaluating modal and amodal segmentations in the ablation studies. Moreover, as objects in common motion cannot be distinguished purely from motion cues, we re-annotate the original DAVIS2017~\cite{Ponttuset17} dataset by grouping jointly moving objects to form a new DAVIS2017-motion dataset. We will release our modified annotations for future research.

\vspace{-0.05cm}
\subsection{Evaluation metrics}
\vspace{-0.1cm}
Following the common practice~\cite{Perazzi16}, we adopt region similarity ($\mathcal{J}$), {\em aka.}~intersection-over-union, 
in single object segmentation tasks. While for multiple object segmentation, we additionally consider the contour similarity ($\mathcal{F}$)~\cite{1273918}.
As for MoCA dataset, since only bounding box annotations are provided, 
we follow the same metric used in~\cite{lin2014microsoft,Yang21a}, and report a mean object detection success rate averaged over different IoU thresholds $\{0.5, 0.6, 0.7, 0.8, 0.9\}$.

\vspace{-0.05cm}
\subsection{Implementation details}
\vspace{-0.1cm}
{\noindent \bf Training details. }
As for pre-processing, we use RAFT to estimate optical flows at $\pm 1$ frame gaps (except for FBMS, where $\pm 3$ frame gaps are adopted to compensate for small
motion), and resize the obtained flows to $128 \times 224$. 
During training, we split the video sequences into $30$ frames per sample, each input frame is first encoded by a U-Net encoder into a feature map with $1/16$ of its original spatial resolution, 
and passed to the transformer bottleneck.
We use $K = 3$ learnable object queries, 
associating to $3$ independent foreground layers.  
The model is trained by the Adam optimizer~\cite{Adam} with a learning rate linearly warmed up to $5\times10^{-5}$ during $40$k iterations, and decreased by half every $80$k iterations.

{\noindent \bf RGB-based test-time adaptation. }
To alleviate the drawbacks from using purely flow information~({\em e.g.}~stationary objects over a long period of time), we investigate the benefits of test-time adaptation with appearance features. For this we use the self-supervised DINO-pretrained vision transformer~\cite{caron2021emerging} (ViT-S/8, patch sizes $8 \times 8$). The model is adapted by using the per-frame object masks obtained from the flow-based model predictions as noisy annotations. These annotations are used to finetune the last two layers of the vision transformer by supervised contrastive learning.

\camera{The finetuned model for each sequence is then applied for mask propagation in the same manner as~\cite{jabri2020walk}. Instead of propagating from the first frame of the sequence,  we formulate a heuristic selection process that picks a single key frame based on temporal coherence between OCLR segmentation predictions. Object masks in this selected frame are then bi-directionally propagated across the whole sequence. Apart from the starting frame information, flow-predicted masks in other frames are also selectively introduced during the mask propagation process as a form of dynamic refinement. Finally, we adopt CRF as a post-processing step to refine the resultant mask predictions. 
More technical details can be found in the Supplementary Material.}

\vspace{-0.05cm}

\subsection{Ablation study}
\vspace{-0.1cm}
Here, we present a series of ablation studies on training details and pipeline for data simulation however, due to the space limitation, we can only summarise some key findings, we refer the reader to Supplementary Materials for all details.

\begin{table}[!htb]
\setlength\tabcolsep{7pt}
\footnotesize
  \caption{\textbf{Settings for training parameters.} ~\textbf{IN}: Instance Normalisation; ~\textbf{Amodal}: Training on amodal mask (vs. modal mask); ~$\mathcal{\lambda}_{\textrm{bound}}$: weight on boundary loss; ~$T$: number of input frames. Syn-Val ($\mathcal{M}$|$\mathcal{A}$) corresponds to modal and amodal results on synthetic dataset.}
  \label{trn-table}
  \centering
  \hspace{-.4cm}
  \begin{tabular}{cccccccc}  
    \toprule
    {} & \multicolumn{4}{c}{Training Settings} & \multicolumn{3}{c}{$\mathcal{J}$ (Mean)  $\uparrow$ }   \\
    \cmidrule(r){2-5}
    \cmidrule(r){6-8}
    Model    & IN     &  Amodal & $\mathcal{\lambda}_{\textrm{bound}}$ & $T$~~~ & Syn-Val ($\mathcal{M}$ | $\mathcal{A}$) & {DAVIS2016} & {\!DAVIS2017-motion\!} \\
    \midrule
    Ours-A & \xmark   &$\checkmark$  & $0.2$  & $30$~~~  &  $83.5$  | $83.0$  & $67.6$ &  $48.7$   \\
    Ours-B & $\checkmark$    & \xmark  & $0.2$  & $30$~~~ &  $81.1$ | $76.9$  &  $69.2$ &  $50.5$     \\ 
    Ours-C & $\checkmark$    &$\checkmark$  & $0.2$  & $30$~~~ &  $\textbf{85.6}$ | $\textbf{84.7}$ & $\textbf{72.1}$ &  $\textbf{54.5}$    \\ 
    \midrule
    Ours-D & $\checkmark$   &$\checkmark$  & $0.2$  & $15$~~~ &  $82.8$ | $83.0$ & $71.3$ & $53.5$    \\
    Ours-E & $\checkmark$    &$\checkmark$  & $0$ & $30$~~~ &  $80.9$ | $81.6$ & $71.5$ & $54.1$    \\ \bottomrule
  \end{tabular}
  \vspace{-0.2cm}
\end{table}

As shown in Table~\ref{trn-table},
we can make the following observations: 
{\em~First}, Instance Normalisation in CNN encoder is indispensable for training the network, as indicated by Ours-A vs.~Ours-C; 
{\em~Second}, 
training on only modal mask degrades model performance on all datasets,
especially on Syn-Val with multiple objects and occlusions,
suggesting that explicit amodal supervision are critical for learning object permanence, as indicated by Ours-B vs.~Ours-D;
{\em~Third}, a longer temporal input~(Ours-C vs.~Ours-D) tends to result in slightly higher overall performance;
{\em~Fourth}, applying boundary loss~(Ours-C vs.~Ours-E) leads to a noticeable performance boost. This validates our assumption that focusing on object boundaries can help the model to learn the information regarding object shapes and layer orders from optical flows.

\vspace{-0.1cm}
\subsection{Single object video segmentation}
\begin{figure}
  \hspace{-0.7cm}
  \includegraphics[width=1.05\textwidth]{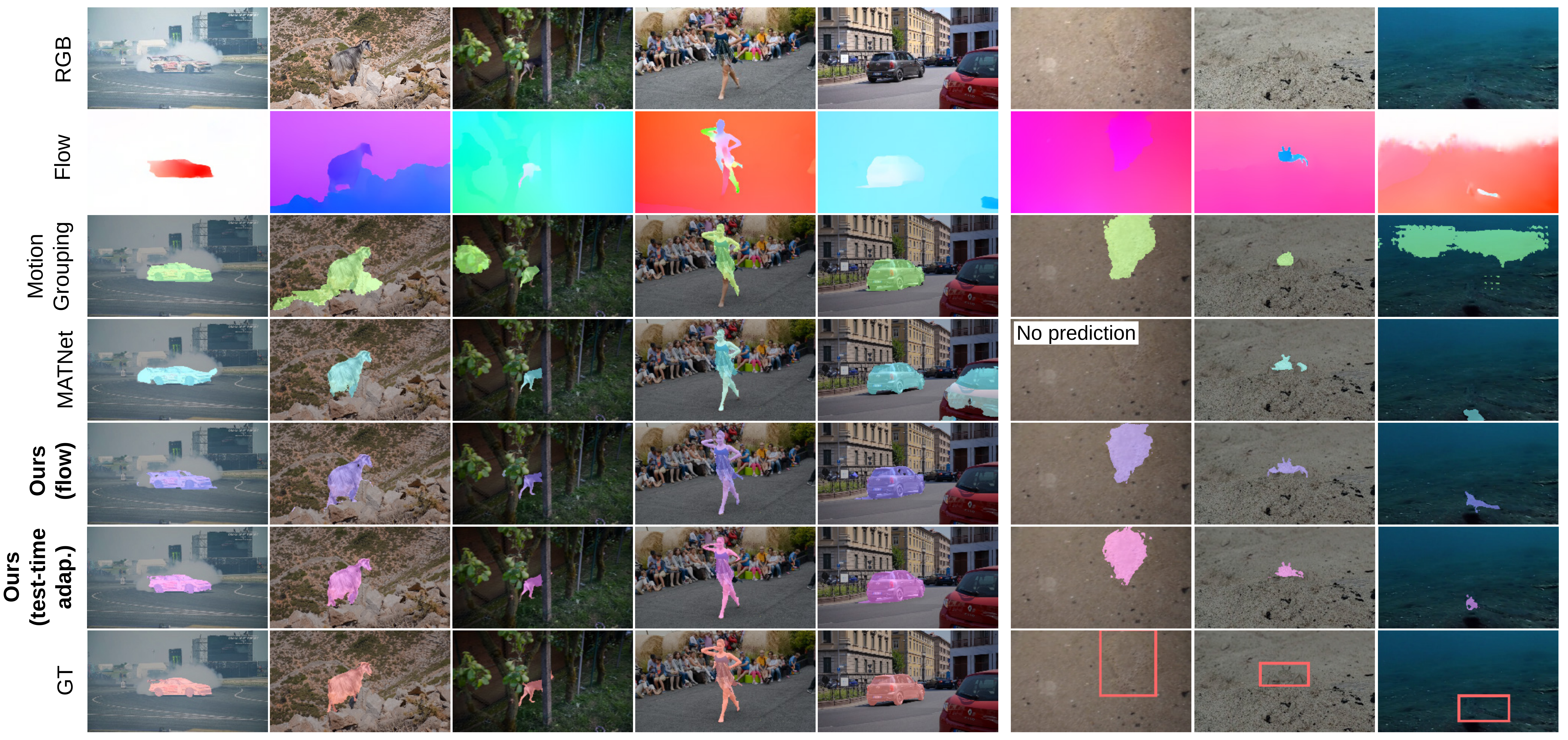}
  \vspace{-0.4cm}
  \caption{Qualitative results of single object video segmentation on DAVIS2016 and MoCA.}
  \label{fig:singleobj2}
  \vspace{-0.4cm}
\end{figure}

\vspace{-0.1cm}
In this section, we compare our model with state-of-the-art methods on various single object segmentation benchmarks.
\revised{Note that, we mainly compare with the self-supervised approaches, as both lines share the same spirit in the sense that training {\em does not} require any manual annotation, nor fine-tuning on the target dataset.}
\revised{As shown in Table~\ref{single1-table}, our flow-only model demonstrates superior performance over all other human-label-free approaches.}
Figure~\ref{fig:singleobj2} provides qualitative illustrations of the model, in comparison with other state-of-the-art methods. 
It can be seen that our predictions are temporally consistent 
and not affected by noticeable background distractors in flow signals. Furthermore, our model is capable of handling complex scenarios including heavy object deformations and occlusions. 
The other methods are not able to maintain the object shape consistently, and consequently their performance is weaker.

\begin{table}[!htbp]
\setlength\tabcolsep{6pt}
  \caption{Quantitative comparison on single object video segmentation benchmarks. \revised{``HA'' stands for human annotations, and   ``SR'' refers to the detection success rate on MoCA. In column Sup. (supervision), ``None'', ``Syn.'', ``Real'' represent self-supervision, synthetic data supervision, and real data supervision, respectively.}  \textbf{\textit{Bold}} represents the state-of-the-art performance (excluding our test-time adaptation results, which are labelled as $\textcolor{blue}{blue}$ instead). For FBMS, results outside/inside brackets correspond to flow inputs with $\pm 1$/$\pm 3$ frame gaps.}
  \label{single1-table}
  \hspace{-.85cm}
  \footnotesize
  \begin{tabular}{ccccccccc}  
    \toprule
    {} & \multicolumn{4}{c}{Training Settings} & \multicolumn{3}{c}{$\mathcal{J}$ (Mean)  $\uparrow$ } & 
    \multicolumn{1}{c}{SR~(Mean)  $\uparrow$ } \\
    \cmidrule(r){2-5}
    \cmidrule(r){6-8}
    \cmidrule(r){9-9}
    Model    & \revised{HA} & \revised{Sup.}  & RGB &  Flow  & DAVIS2016 & SegTrackv2 & FBMS-59 & MoCA\\
    \midrule
    NLC~\cite{faktor2014videonlc} &\revised{\xmark} & \revised{None}  & $\checkmark$  &$\checkmark$  &  $55.1$ &  $67.2$  &  $51.5$ &  $-$   \\
    CIS (w.~post-process.)~\cite{yang_loquercio_2019} & \revised{\xmark} & \revised{None}  & $\checkmark$  &$\checkmark$  &  $71.5$ &  $62.0$  &  $63.5$  & $0.363$   \\
    Motion Grouping~\cite{Yang21a}& \revised{\xmark} & \revised{None}  & \xmark  &$\checkmark$  &  $68.3$ &  $58.6$  &  $53.1$  &  $0.484$ \\

    SIMO~\cite{Lamdouar21} & \revised{\xmark} & \revised{Syn.}  & \xmark  &$\checkmark$  &  $67.8$ &  $62.0$  &  $-$  & $0.566$   \\
    \textbf{OCLR (flow-only)}& \revised{\xmark} & \revised{Syn.}  & \xmark  &$\checkmark$  &  $\textbf{72.1}$ &  $\textbf{67.6}$  &  $\textbf{65.4 (70.0)}$ &   $\textbf{0.599}$  \\
    \camera{\textbf{OCLR (test. adap.)}} & \revised{\xmark} & \revised{Syn.}  &$\checkmark$  &$\checkmark$  &  \camera{$\textcolor{blue}{80.9}$} &  \camera{$\textcolor{blue}{72.3}$}  &  $\textcolor{blue}{69.8}$ \textcolor{blue}{(}$\textcolor{blue}{72.7}$\textcolor{blue}{)} &  \camera{$0.559$} \\
    \midrule
    FSEG~\cite{Jain17} & \revised{$\checkmark$} & \revised{Real}  & $\checkmark$  & $\checkmark$  &  $70.7$ &  $\textbf{61.4}$  &  $68.4$ &  $-$
      \\
    COSNet~\cite{Lu_2019_CVPR} & \revised{$\checkmark$}  & \revised{Real}  & $\checkmark$  & \xmark  &  $80.5$ &  $49.7$   &  $75.6$  &  $0.417$  \\
    MATNet~\cite{zhou20} & \revised{$\checkmark$}  & \revised{Real}  & $\checkmark$  &$\checkmark$  &  $82.4$  &  $50.4$   &  $\textbf{76.1}$  &  $\textbf{0.544}$   \\
    D$^2$Conv3D~\cite{D2conv3d} & \revised{$\checkmark$} & \revised{Real}   & $\checkmark$  & \xmark  &  $\textbf{85.5}$  &  $-$   &  $-$  &   $-$  \\
    \bottomrule
  \end{tabular}
  \vspace{-0.25cm}
\end{table}

The RGB-based test-time adaptation gives a further performance boost on most of the benchmarks, 
and is even competitive to some supervised approaches that have been finetuned on the target video data. Note that the test-time adaptation is actually detrimental to performance on MoCA. This is not unexpected though, as MoCA has many camouflage sequences where the objects are not visually distinguishable in appearance from their background environment, and motion provides crucial cues for discovering them.

\vspace{-0.1cm}
\subsection{Multiple object video segmentation}
\vspace{-0.1cm}
To the best of our knowledge, 
no existing unsupervised approaches have reported performance segmenting multiple objects purely based on optical flow.
\revised{Apart from re-running the original self-supervised Motion Grouping~\cite{Yang21a} method (with three foreground queries) as a baseline on DAVIS2017-motion, we also train a variant of the Motion Grouping model directly supervised by our synthetic dataset, namely Motion Grouping (sup.).} \camera{As an additional baseline, we train a Mask R-CNN~\cite{matterport_maskrcnn_2017} model with only optical flow inputs on our synthetic data, and refer to this as Mask R-CNN (flow-only).}
We also compare with existing semi-supervised approaches,
where the first-frame ground-truth segmentation is provided,
and the model only needs to propagate the masks through the video,
thus an easier task than ours which requires to simultaneously discover the objects and track them.

\begin{figure}
  \hspace{-0.9cm}
  \includegraphics[width=1.05\textwidth]{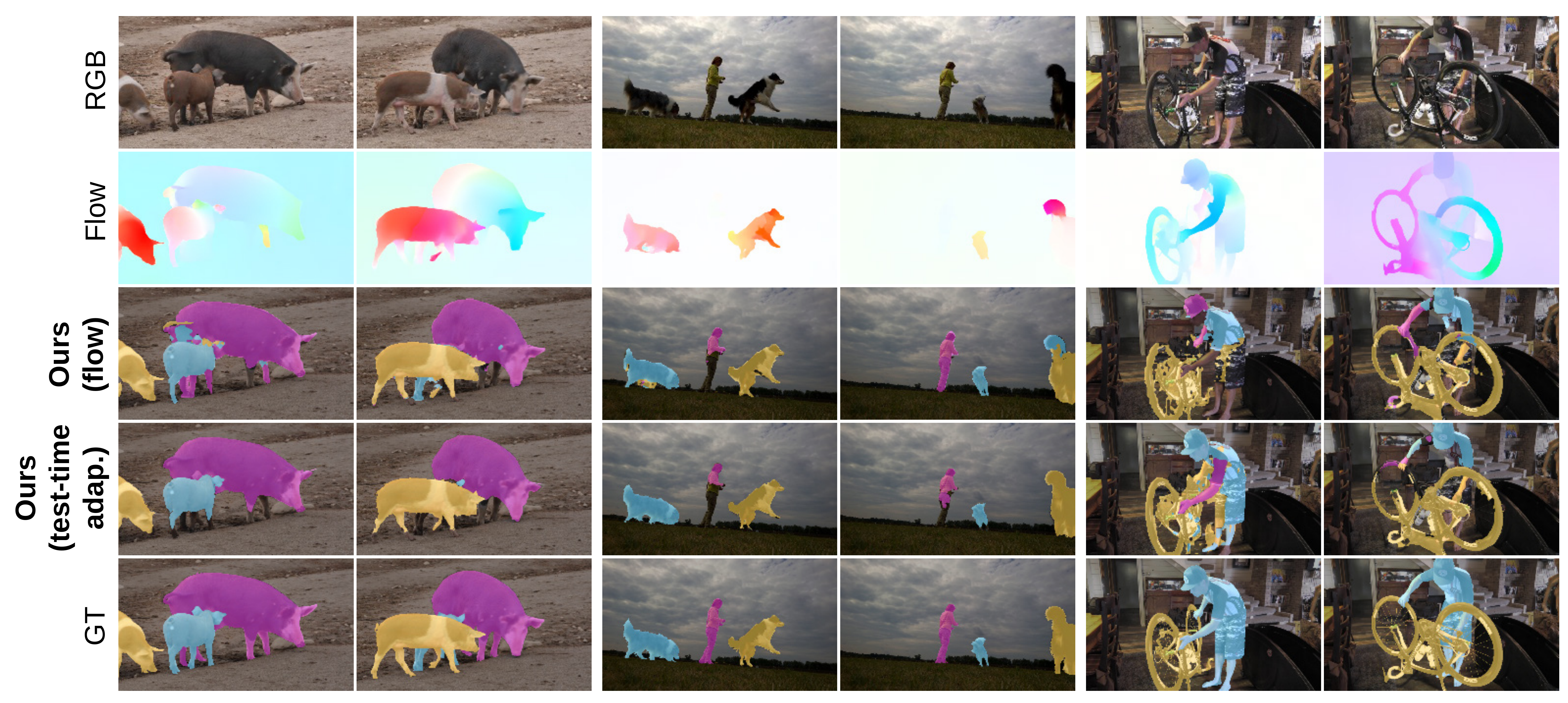}
  \caption{Qualitative results of multi-object video segmentation on DAVIS2017-motion.}
  \label{fig:multiobj}
  \vspace{-0.55cm}
\end{figure}

\camera{As shown in Table~\ref{multi-table}, our proposed flow-only model outperforms both the Motion Grouping and Mask R-CNN baselines.}
This is because our OCLR model exploits a layered representation to maintain the object shape through the video, which enables to segment the objects that are under occlusion or having unnoticeable motion.
In contrast, Motion Grouping only predicts per-frame segmentations,{\em~i.e.} not forced to preserve temporal relations between objects. 
This conjecture can also be confirmed by the qualitative results in Figure~\ref{fig:multiobj}, for example, despite the person and one dog (4th column) are not visible in the flow, our model still correctly recovers them.
Moreover, after RGB-based test-time adaptations, 
the performance can be further boosted both quantitatively and qualitatively, for example, the boundary of the pig mask in 1st column, the person in the 5th column.

\begin{table}[!htb]
\footnotesize
\setlength\tabcolsep{9pt}
  \caption{Quantitative comparison of multi-object video segmentation on DAVIS2017-motion. 
  \camera{Note that, the compared methods here are trained without using any human annotations during training, in particular, Motion Grouping (sup.), Mask R-CNN (flow-only) and OCLR models are supervised by only synthetic data, and other approaches are trained with self-supervision.}
  \textbf{\textit{Bold}} represents the state-of-the-art performance (excluding our test-time adaptation results, which are labelled as $\textcolor{blue}{blue}$ instead).}
  \label{multi-table}
  \hspace{-.45cm}
  \begin{tabular}{ccccccc}  
    \toprule
    {} & \multicolumn{3}{c}{Training settings} & \multicolumn{3}{c}{DAVIS2017-motion performance } \\
    \cmidrule(r){2-4}
    \cmidrule(r){5-7}
    Model    & {1st-frame-GT} & RGB &  Flow  & {$\mathcal{J}$\&$\mathcal{F}$  $\uparrow$} & {$\mathcal{J}$ (Mean)  $\uparrow$} & {$\mathcal{F}$ (Mean)  $\uparrow$}\\
    \midrule
    Motion Grouping~\cite{Yang21a} & \xmark & \xmark  &$\checkmark$  &  $35.8$ & $38.4$  &  $33.2$    \\
    \revised{Motion Grouping (sup.)} & \revised{\xmark} & \revised{\xmark}  & \revised{$\checkmark$}  &  \revised{$39.5$} & \revised{$44.9$}  &  \revised{$34.2$}    \\
    \camera{Mask R-CNN (flow-only)} & \camera{\xmark} & \camera{\xmark}  & \camera{$\checkmark$}  &  \camera{$50.3$} & \camera{$50.4$}  &  \camera{$50.2$}    \\
    \textbf{OCLR (flow-only)} &  \xmark  &  \xmark  &$\checkmark$  &  $\textbf{55.1}$ &  $\textbf{54.5}$  &  $\textbf{55.7}$    \\
    \camera{\textbf{OCLR (test. adap.)}} & \xmark  & $\checkmark$  & $\checkmark$  &  $\textcolor{blue}{64.4}$ &  $\textcolor{blue}{65.2}$  &  $\textcolor{blue}{63.6}$  \\
    \midrule
    CorrFlow~\cite{Lai19} & $\checkmark$  & $\checkmark$  &  \xmark &  $54.0$ &  $54.2$  &  $53.7$     \\
    UVC~\cite{nips19_joint_task} & $\checkmark$  & $\checkmark$  &  \xmark &  $65.5$ &  $66.2$  &  $64.7$ \\
    MAST~\cite{Lai20} & $\checkmark$  & $\checkmark$  &  \xmark &  $70.9$ &  $71.0$  &  $70.8$    \\
    CRW~\cite{jabri2020walk} & $\checkmark$  & $\checkmark$  & \xmark &  $73.4$ &  $72.9$  &  $74.1$     \\
    DINO~\cite{caron2021emerging} & $\checkmark$  & $\checkmark$  &  \xmark  & $\textbf{78.7}$ &  $\textbf{77.7}$  &  $\textbf{79.6}$     \\
    \bottomrule
  \end{tabular}
  \vspace{-.45cm}
\end{table}

\section{Related Work}
\vspace{-0.15cm}
\label{sec:related}
\par{\noindent \textbf{Video object segmentation}}
has been a longstanding task in computer vision,
two protocols have attracted increasing interest in the recent literature~\cite{Brox10,Ochs11,Fragkiadaki12,Papazoglou13,Keuper15,Bideau16a,Jun17,Jain17,cvpr17_OSVOS,Ponttuset17,Xu18-YTB,Tokmakov19,Dave19,iccv19_stm,cvpr19_feelvos,Vondrick18,tpami18_osvos-s,Fan19,Yang19,Lai19,Lai20},
namely, semi-supervised video object segmentation~({\bf semi-supervised VOS}),
and unsupervised video object segmentation~({\bf unsupervised VOS}).
The former aims to re-localize one or multiple targets that are specified in the first frame of a video with pixel-wise masks, 
and the latter considers automatically segmenting the object of interest~(usually the most salient one) from the background in a video sequence.
Despite being called {\bf unsupervised VOS}, in practice,
the popular methods address such problems by supervised training on large-scale external datasets, this is in contrast to our proposed approach that does not rely on human annotations whatsoever. 
\vspace{-0.05cm}
\par{\noindent \textbf{Motion segmentation}}
focuses on discovering the \emph{moving} objects in videos.
In the literature,
\cite{Brox10,Ochs11,Keuper15,xie19} proposed to cluster the pixels with similar motion patterns;
\cite{Tokmakov17,Tokmakov19,Dave19} train deep networks to map the motions to segmentation masks.
In \cite{yang_loquercio_2019},
adversarial training was adopted to leverage the independent motions between the moving object and its background;
In~\cite{Bideau16,Bideau18,Lamdouar20}, 
the authors propose to highlight the independently moving object by compensating for the background motion, either by registering consecutive frames, or explicitly estimating camera motion.
In~\cite{Yang21a}, a Transformer-like architecture is used to reconstruct the input flows, and the segmentation masks can be generated as a side product. 
In contrast to the existing approaches that primarily focus on single moving object segmentation, we focus on segmenting \textbf{multiple} moving objects, even under mutual occlusions.

\vspace{-0.05cm}
\par{\noindent \textbf{Layered representation}} 
was originally proposed by Wang and Adelson~\cite{Wang94},
to represent a video as a composition of layers with simpler motions.
Recently, the layer decomposition ideas have been adopted for novel view synthesis~\cite{Zhou18,Srinivasan19}, 
separating reflections and other semi-transparent effects~\cite{Alayrac19a,Alayrac19b,Gandelsman19,lu20,lu21}, 
or foreground/background estimation~\cite{Gandelsman19}.
Unlike these approaches that primarily focus on graphics applications,
we propose a layered representation to handle the occlusion problems in multi-object discovery purely from their motions. 

\vspace{-0.05cm}
\par{\noindent \textbf{Object-centric representation}}
decomposes the scenes into ``objects'',
normally, visual frame reconstruction has been widely used as training objective, for example, IODINE~\cite{Greff19} uses iterative variational inference to infer a set of latent variables recurrently, 
with each representing one object in an image.
Similarly, MONet~\cite{Burgess19} and GENESIS~\cite{Engelcke20} also adopt multiple encoding-decoding steps,
\cite{locatello2020object,Kipf22} propose a slot attention mechanism, which enables the iterative binding procedure.
In this paper, we also adopt a Transformer variant, 
but focus on discovering objects in videos by motions.

\par{\noindent \camera{\textbf{Amodal segmentation}}} 
\camera{refers to segmenting the complete shape of objects, 
including the invisible parts due to possible occlusions. 
Some existing approaches address this problem by applying human-estimated~\cite{8954364, yuting2021amodal, sun2021amodal} or synthetic~\cite{8953603} supervision, while other works generate training datasets with synthetic occluders~\cite{Lieccv2016, zhan2020self, NEURIPS2020_bacadc62}. Although not specifically targeting the amodal segmentation task, our work leverages the idea of amodal perception and grants the trained model a notion of object permanence via synthetic amodal mask supervision.}

\section{Discussion}
\label{sec:discussion}
\vspace{-0.2cm}
We have achieved our design goal, in that the model is able to handle multiple objects, and their occlusions, and to predict depth ordering of their layers. 
\revised{Also, the model demonstrates superior performance, both quantitatively and qualitatively, over prior methods that rely on zero human annotations.}
Nevertheless, there are of course some limitations and room for further improvements: 
{\em~First}, the current method may fail in very challenging real-world scenarios such as heavy object deformations, complex mutual interactions, etc. Probably a more sophisticated data simulation pipeline with highly articulated objects or even 3D sprites would help to further reduce this Sim2Real domain gap.
{\em~Second}, our test-time adaptation has demonstrated a remarkable performance improvement benefiting from the combination between flow inferences and RGB correlations. This overcomes, to an extent, the problem of the flow-only model `forgetting' the shape of objects when they are stationary for multiple frames. A further direction could naturally be to incorporate RGB information into our flow-based network in pre-training.
\revised{{\em~Third}, the model currently uses a global depth ordering for the sequence, so it cannot handle situations such as order altering and mutual occlusions. Potential future studies could work on a more sophisticated layered model by focusing on pair-wise relationships between objects.}

\vspace{-0.05cm}
Despite these limitations, the approach has convincingly demonstrated both the value of inferring amodal segmentation masks, in order to handle occlusions,  and the possibility of training such models entirely on synthetic sequences.

\camera{\subsection*{Acknowledgements}
This research is funded by EPSRC Programme Grant VisualAI EP/T028572/1, a Royal Society Research Professorship RP\textbackslash R1\textbackslash 191132, and a Clarendon Scholarship. We thank Tengda Han for proof-reading.}

{\small
\bibliographystyle{ieee_fullname}
\bibliography{egbib}
}

\vspace{-1.0cm}

\newpage
\appendix
{
\hypersetup{linkcolor=black}
\mtcsettitle{parttoc}{}
\mtcsetrules{*}{off}
\setlength{\mtcindent}{-1.5em}
\doparttoc %
\faketableofcontents
\part{\large{Supplementary material}} %
\parttoc %
}
\section{Architecture and implementation details}
In this section, we describe the architecture of the OCLR model in detail.

\subsection{CNN backbone}
As shown in Figure~\ref{fig:arch}, we adopt a U-Net architecture to extract
the visual features for each input frame, and to upsample the feature map output from the transformer-based bottleneck.
The details of the U-Net backbone are provided in Table~\ref{tab:cnn}.

\begin{table}[!htbp]
\setlength\tabcolsep{6pt}
  \caption[U-Net architecture as the CNN backbone]{\textbf{U-Net architecture.} The U-Net uses
  a CNN as the backbone. All convolution operations are followed by an Instance Normalisation and a ReLU activation, except for the "de-output" stage in the U-Net decoder.}
  \label{tab:cnn}
  \hspace{-0.7cm}
  \footnotesize
  \begin{tabular}{ccc|ccc}  
    \toprule
    \multicolumn{3}{c}{Encoder} & \multicolumn{3}{c}{Decoder} \\
    \cmidrule(r){1-3}
    \cmidrule(r){4-6}
    stage  & operation  & output size  &  stage  & operation  & output size \\
    \midrule
     en-input  & $-$  & $2\times128\times224$  &  de-input  & $-$  & $512\times8\times14$     \\
    \midrule
     en-conv1  & ($3\times3, 32$) $\times 2$ & $32\times128\times224$  &  de-conv$^T$4  & stride $=2$, 256  & $256\times16\times28$     \\
     en-mp1 & maxpool, stride $=2$ & $32\times64\times112$ & de-sc4 & en-conv4 skip connect. & $512\times16\times28$ \\
     $-$  & $-$ & $-$  &  de-conv4  & ($3\times3, 256$) $\times 2$ & $256\times16\times28$     \\
     en-conv2  & ($3\times3, 64$) $\times 2$ & $64\times64\times112$  &  de-conv$^T$3  & stride $=2$, 128  & $128\times32\times56$        \\
     en-mp2 & maxpool, stride $=2$ & $64\times32\times56$ & de-sc3 & en-conv3 skip connect. & $256\times32\times56$ \\ 
     $-$  & $-$ & $-$  &  de-conv3  & ($3\times3, 128$) $\times 2$ & $128\times32\times56$     \\
     en-conv3  & ($3\times3, 128$) $\times 2$ & $128\times32\times56$  &  de-conv$^T$2  & stride $=2$, 64  & $64\times64\times112$      \\
     en-mp3 & maxpool, stride $=2$ & $128\times16\times28$ & de-sc2 & en-conv2 skip connect.  & $128\times64\times112$  \\ 
     $-$  & $-$ & $-$  &  de-conv2  & ($3\times3, 64$) $\times 2$ & $64\times64\times112$     \\
     en-conv4  & ($3\times3, 256$) $\times 2$ & $256\times16\times28$ &  de-conv$^T$1  & stride $=2$, 32  & $32\times128\times224$       \\
     en-mp4 & maxpool, stride $=2$ & $256\times8\times14$ &de-sc1 &en-conv1 skip connect.  & $64\times128\times224$  \\ 
     $-$  & $-$ & $-$  &  de-conv1  & ($3\times3, 32$) $\times 2$ & $32\times128\times224$     \\
     en-bottleneck  & ($3\times3, 512$) $\times 2$ & $512\times8\times14$  &  de-output  & $3\times3, 3$ & $3\times128\times224$     \\
    \bottomrule
  \end{tabular}
  \vspace{-0.2cm}
\end{table}

\subsection{Transformer-based bottleneck}
\begin{figure}[!htbp]
  \includegraphics[width=1.1\textwidth]{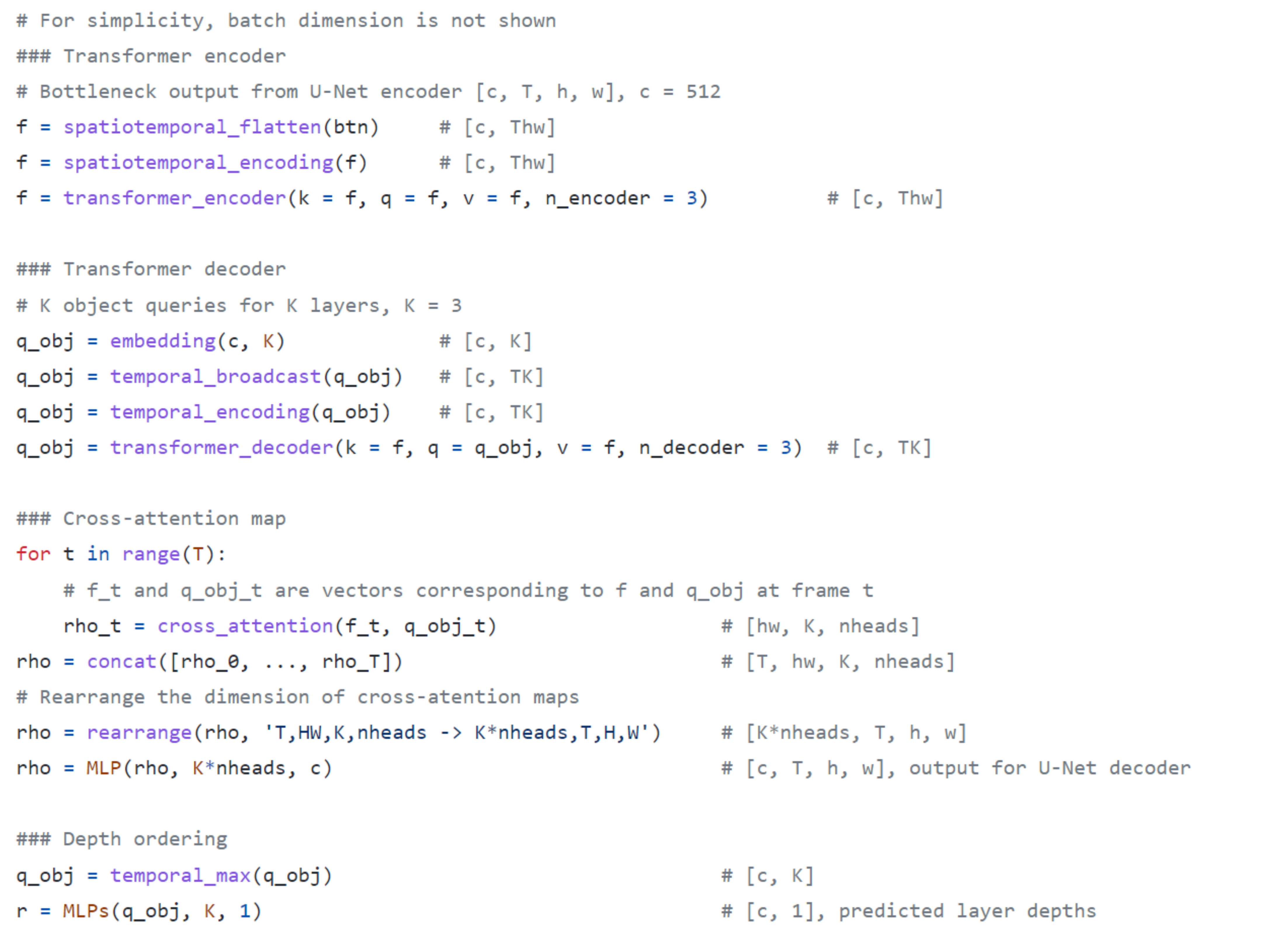}
  \caption[Pseudo-code for Transformer-based bottleneck]{Pseudo-code for Transformer-based bottleneck.}
  \label{supfig:transbtn}
  \hspace{-1.7cm}
\end{figure}

In Figure~\ref{supfig:transbtn}, 
we provide pseudo-code for obtaining cross-attention feature maps and the layer depth order in the Transformer-based bottleneck.

\subsection{Training details}
The model is trained on the synthetic dataset from about $120$k frames. The input batch size is set to $2$, 
and each input sequence has $30$ frames (optical flow). 
During training, we apply a boundary loss weight $\lambda_{\text{bound}} = 0.2$ and an ordering loss weight $\lambda_{\text{order}} = 0.05$. The learning rate is linearly warmed up to $5\times10^{-5}$ during the initial $40$k iterations, followed by $50\%$ decays every $80$k iterations. 
All the models are trained on a single NVIDIA Tesla V100 GPU with $32$G memory, and the full convergence takes approximately $5$ days with $600$k iterations.

\section{Synthetic data generation pipeline}
In this section, we elaborate on our synthetic data generation process, with more implementation details on background and foreground objects, 
dataset distribution, together with several examples.

\begin{figure}[t]
  \hspace{-0.7cm}
  \includegraphics[width=1.05\textwidth]{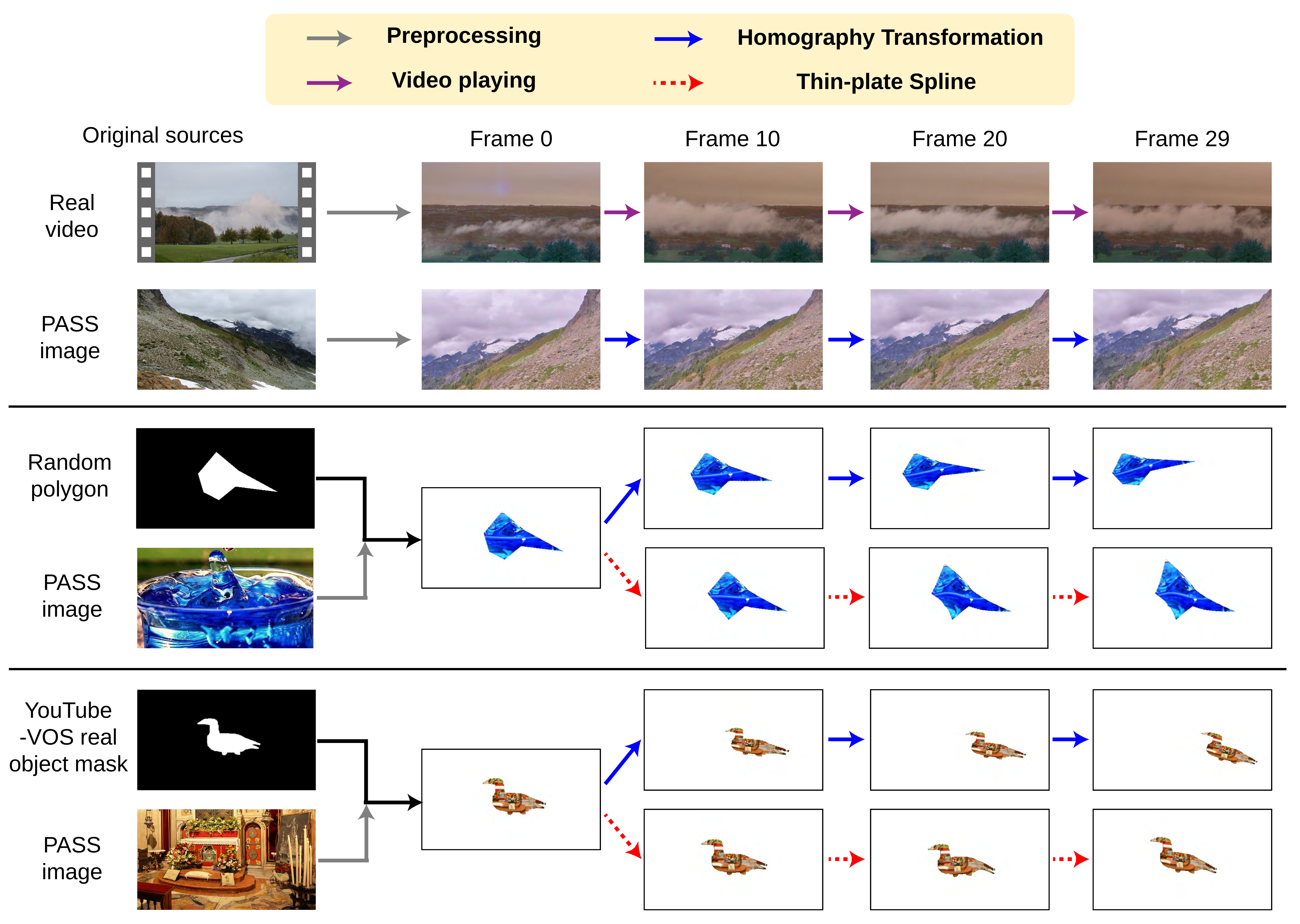}
  \caption[Synthetic data generation]{Synthetic data generation.}
  \label{supfig:syn_gen}
\end{figure}

\subsection{Background}
\paragraph{Real background video.} 
As shown in Figure~\ref{supfig:syn_gen}, 
we select real-world background videos from copy-right-free sources, 
followed by frame-wise pre-processing including random cropping, colour jittering, flipping and reflections. 
Augmented video frames are then sampled as backgrounds, 
in a chronological or reverse chronological order, 
starting from a random frame with several different frame rates. 

\paragraph{Homography-transformed image.} 
Apart from using real videos, we generate background frames by randomly sampling images from the PASS dataset~\cite{Asano21}. 
After a similar pre-processing to that for videos above, 
the augmented image is set as frame 0. 
Backgrounds for later frames are propagated by applying homography transformations to simulate the camera motion.

\subsection{Foreground objects}
\paragraph{Shapes and textures.} 
As shown in  Figure~\ref{supfig:syn_gen},
the foreground objects are obtained from two sources,
one is by generating polygon shapes by connecting $3$ to $8$ points with random coordinates as vertices, and the other is by taking  real object masks from the YouTube-VOS 2018 training set, followed by random resizing and rotation. Both object shapes are textured by pre-processed PASS images to form object sprites. 

\paragraph{Foreground object motion simulation.}
At frame 0, the sprites are initialised at random positions. 
(In Figure~\ref{supfig:syn_gen}, they are initialised in the middle of the frame for demonstration purposes.) We further apply two transformations to simulate object motion, 
namely homography transformations and thin-plate spline mappings. 
Homographies include rotations, scaling, perspective distortions and spatial translations. 
Thin-plate splines transform coordinates according to the motion of control points and include elastic-like stretching. 
For the examples shown in Figure~\ref{supfig:syn_gen}: for the polygon sprite, 
the top and bottom left vertices are moving control points; 
while for the real duck-shaped object, the moving point is at the ``tail'' of the duck.

\paragraph{Stationary object.} 
In Figure~\ref{supfig:syn_demo}, the first sequence (from the top) illustrates an example of a stationary object. Between frames $t_1$ and $t_2$, the object in layer 2 has no relative motion with respect to the background. 
As a result, it disappears in the flow field at frame $t_1$. 
In practise, stationary objects are achieved by matching object transformations to background motions (therefore only applicable for homography backgrounds).

\subsection{Layer composition}
\begin{figure}[!htb]
  \hspace{-0.7cm}
  \includegraphics[width=1.05\textwidth]{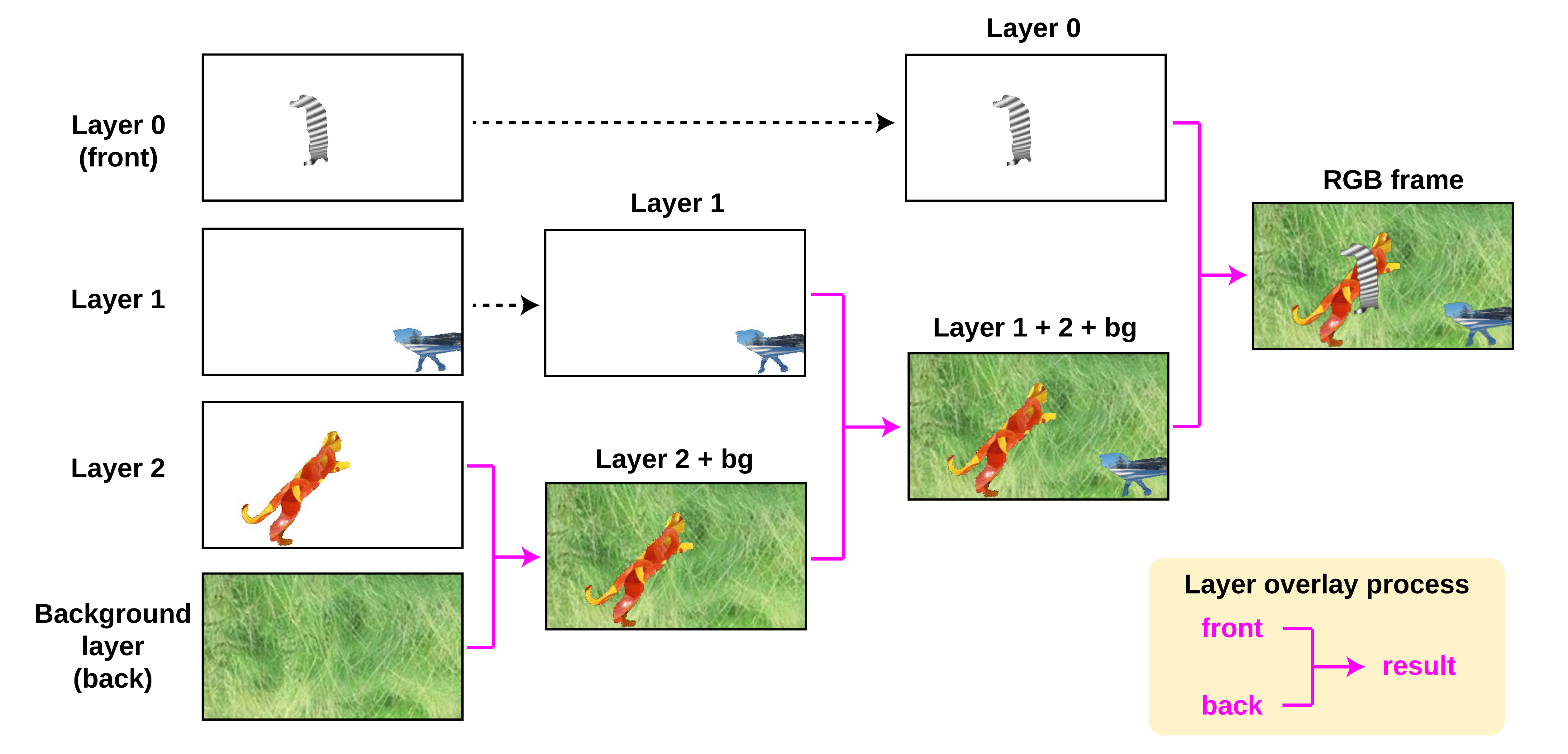}
  \caption[Layer composition]{\textbf{Layer composition from back to front.}  The order from front to back: Layer 0, Layer 1, Layer 2, Background layer.}
  \label{supfig:syn_compose}
\end{figure}

\begin{figure}[!htb]
  \hspace{-1.0cm}
  \includegraphics[width=1.1\textwidth]{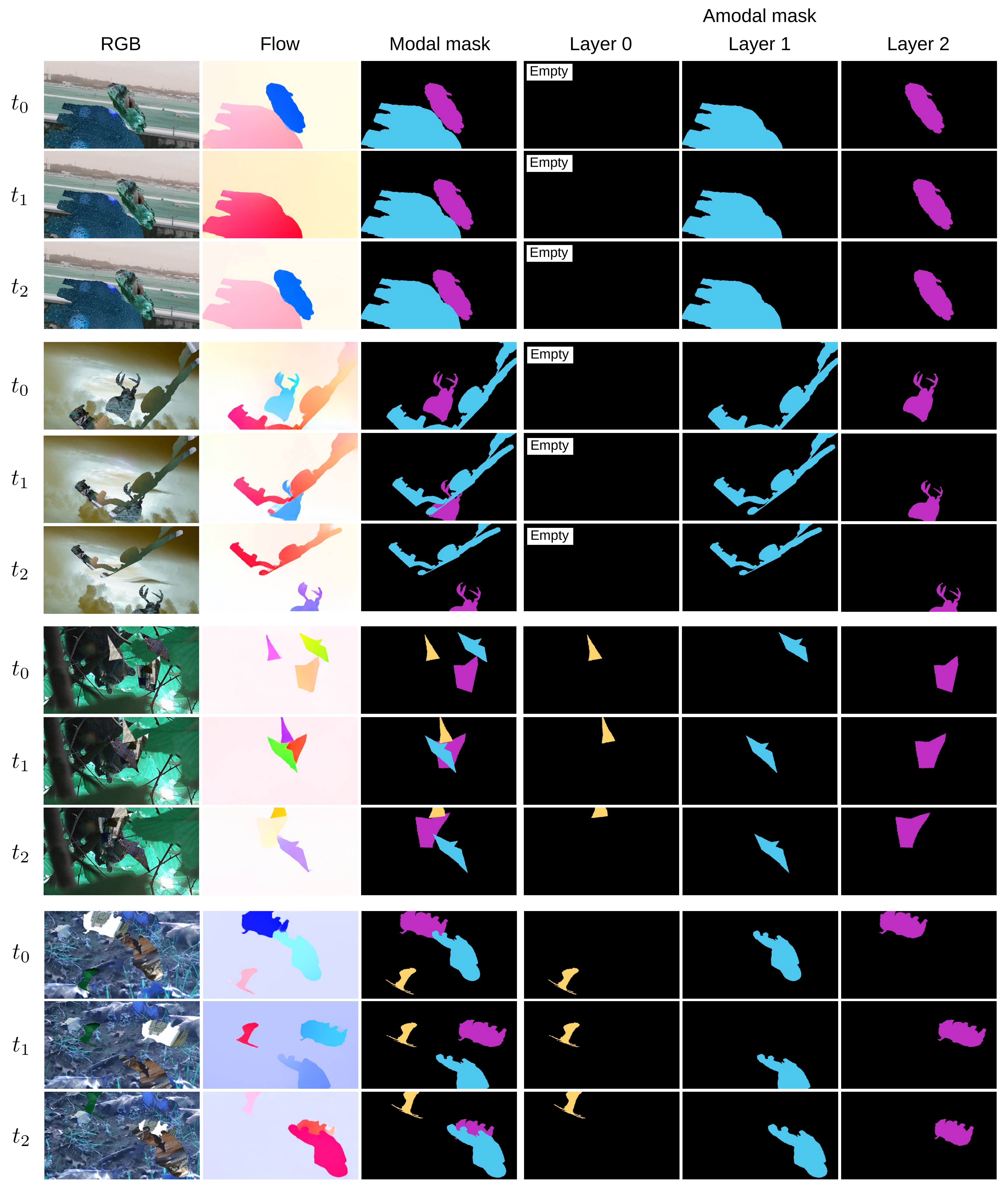}
  \caption[Examples of synthetic data with ground-truth annotations]{\textbf{Examples of synthetic data with ground-truth annotations.} From top to bottom: 2-objs sequence with stationary objects, 2-objs sequence with occlusions, 3-objs sequence with polygon sprites, 3-objs sequence with real object sprites. For demonstration purposes, the first sequence (from the top) shows three \textbf{consecutive} frames, whereas for the rest sequences, three frames are selected at fixed time intervals. }
  \vspace{1.5cm}
  \label{supfig:syn_demo}
\end{figure}

In the previous sections, 
we have explained how objects and backgrounds can be simulated independently. The synthetic video sequence is obtained by a layer composition process that combines all the introduced components via a back-to-front blending process as shown in Figure~\ref{supfig:syn_compose}. Examples of composited sequences are given in Figure~\ref{supfig:syn_demo}. We also refer an animated layer composition demonstration and more example sequences to the supplementary video.

\subsection{Dataset distribution}
In Table~\ref{suptab:syn_dis}, we provide the details of the video datasets generated by the simulation pipeline. 
Overall, there are three major pipeline settings: background types, object types and object motions. 
For all synthetic datasets, 
an equal proportion of homography transformed backgrounds and real video backgrounds are generated. 
Similarly, we evenly split sequences with polygon and real object sprites. Regarding object motion simulations, we emphasize the object non-rigidness by applying a combination of homography transformations and thin-plate splines to most sequences (around two times more than sequences with only homography transformations.) Moreover, we randomly select one-third of all sequences and introduce stationary objects for $1$ to $5$ frames.

On the dataset splits: \textbf{Syn-train} is used as the main dataset to train the OCLR flow-based model, consisting of $4608$ sequences with over $138$k frames. \textbf{Syn-single} and \textbf{Syn-Val} are mainly used for validation in ablation studies. The former contains $256$ single-object-only sequences with a total $7.7$k frames, 
and is treated as a simplified dataset used for tuning synthetic pipeline settings. On the other hand, Syn-Val consists of $384$ multi-object sequences~(around $12$k frames) with $1,2,3$ objects in equal proportions. It is mainly used for validating multi-object segmentation performance, 
with both modal and amodal results reported. 

\begin{table}[!htbp]
\setlength\tabcolsep{5pt}
  \caption[Synthetic dataset distribution]{\textbf{Synthetic dataset distribution.} \textbf{homo-bg:} homography transformed image background; \textbf{real-bg:} real video background; \textbf{homo:} homography transformations only; \textbf{homo+tps:} homography transformations and thin-plate spline; \textbf{polygon:} polygon sprites; \textbf{real-obj:} real object sprites. Unless specified otherwise, all values in the table denote the number of sequences (each with $30$ frames).}
  \label{suptab:syn_dis}
  \hspace{-1.0cm}
  \footnotesize
  \begin{tabular}{cccccccccccc}  
    \toprule
    \multicolumn{3}{c}{Pipeline settings} & \multicolumn{4}{c}{Syn-Train} & {Syn-Single} & \multicolumn{4}{c}{Syn-Val} \\
    \cmidrule(r){1-3}
    \cmidrule(r){4-7}
    \cmidrule(r){8-8}
    \cmidrule(r){9-12}
    Background & Obj. motion  & Obj. type  &  1-obj  & 2-objs  & 3-objs & Total & 1-obj & 1-obj  & 2-objs  & 3-objs & Total  \\
    \midrule
     homo-bg & homo  & polygon  &  $96$  & $96$  & $96$ & $288$   & $16$ & $8$ & $8$ & $8$ & $24$ \\
     homo-bg & homo  & real-obj  &  $96$  & $96$  & $96$ & $288$  & $16$ & $8$ & $8$ & $8$ & $24$\\
     homo-bg & homo+tps  & polygon  &  $288$  & $288$  & $288$ & $864$  & $48$ & $24$ & $24$ & $24$ & $72$\\
     homo-bg & homo+tps  & real-obj  &  $288$  & $288$  & $288$ & $864$  & $48$ & $24$ & $24$ & $24$ & $72$\\
     real-bg & homo  & polygon  &  $96$  & $96$  & $96$ & $288$   & $16$ & $8$ & $8$ & $8$ & $24$\\
     real-bg & homo  & real-obj  &  $96$  & $96$  & $96$ & $288$   & $16$ & $8$ & $8$ & $8$ & $24$ \\
     real-bg & homo+tps  & polygon  &  $288$  & $288$  & $288$ & $864$  & $48$ & $24$ & $24$ & $24$ & $72$ \\
     real-bg & homo+tps  & real-obj  &  $288$  & $288$  & $288$ & $864$  & $48$ & $24$ & $24$ & $24$ & $72$\\
     \midrule
     \multicolumn{3}{c}{Total number of sequences} &   $1536$  & $1536$ & $1536$  & $4608$ & $256$ & $128$ & $128$ & $128$  & $384$\\
     \midrule
     \multicolumn{3}{c}{Total number of frames} &   $46.8$k  & $46.8$k & $46.8$k & $138.2$k & $7.7$k & $3.8$k & $3.8$k & $3.8$k & $11.5$k \\
    \bottomrule
  \end{tabular}
  \vspace{-0.2cm}
\end{table}

\section{RGB-based test-time adaptation}
In this section, 
we detail the procedure for applying RGB-based sequences as test-time adaptation.
Specifically, 
given the predicted modal segmentations from our \textbf{OCLR~(flow-only)} model,  we can adopt a similar mask propagation process to that in self-supervised tracking~\cite{Vondrick18, Lai20, jabri2020walk}.
Overall, the test-time adaptation process proceeds in three stages: first, finetuning a DINO-pretrained vision transformer (ViT)~\cite{caron2021emerging}; second, mask selection and propagation; and, third, dynamic refinement during mask propagations.

\paragraph{Finetuning of DINO-pretrained vision transformer.} 
Given a video sequence with $T$ RGB frames, 
$\mathcal{V}_{\text{RGB}} = \{I_1, \dots, I_T\}$,
$I_t \in \mathbb{R}^{480 \times 832 \times 3}$,
we can compute their RGB features with a pre-trained self-supervised vision transformer, namely, DINO-ViT-S/8~\cite{caron2021emerging}~(patch sizes $8 \times 8$).

\begin{equation}
    \mathcal{F}_{\text{RGB}}=\{f_1, \dots, f_T\} = \{\Phi_{\text{DINO}}(I_1), \dots, \Phi_{\text{DINO}}(I_T)\}
\end{equation}
where $f_t \in \mathbb{R}^{60 \times 104 \times 384}$.

In order to adapt the DINO model for our purpose, 
we use the predicted modal masks from our \textbf{OCLR~(flow-based) model} as noisy annotations to finetune the last two layers of the vision transformer (ViT) by noise contrastive estimation~(NCE).
In detail, for each object mask in frame $t$, we define a tri-map with positive $P_t$, negative $N_t$ and uncertain $U_t$ regions, where uncertain regions are normally a $5$-pixel wide exterior to the object mask. 

Considering two frames $t$ and $t + n$ ($n \in [1,4]$), 
for each pixel in the positive region in one frame, {\em e.g.}$f_{t,j}$~($j \in P_t$), 
we can compute its cosine similarities to all features in the positive region of $P_{t+n}$, 
and treat this as the positive score:
\begin{equation}
PS_j = \sum_{k \in P_{t+n}}{f_{t,j} \cdot f_{t+n,k}}
\end{equation}
Similarly, the negative sample is defined as
\begin{equation}
NS_j = \sum_{k' \in N_{t+n}}{f_{t,j} \cdot f_{t+n,k'}}
\end{equation}
The process is repeated for all pixels in $P_t$ region to give an averaged InfoNCE loss
\begin{equation}
\mathcal{L}_{\text{NCE}}= -\frac{1}{|P_{t}|}\sum_{j \in P_{t}}{\log{\frac{PS_j}{PS_j + NS_j}}}
\end{equation}
For each test sequence, we apply this contrastive loss to finetune the DINO-pretrained ViT by an Adam optimizer with a learning rate of $1\times10^{-5}$ that linearly decays over $1$k iterations.

\paragraph{Mask selection and propagation.} 
After finetuning the DINO model, 
we can use it to propagate the predicted segmentation masks from our \textbf{OCLR~(flow only)} model. 
Specifically, the predictions for each pixel at frame $t$ can be obtained by computing its nearest neighbour pixels in previous frames, and copying their labels. 
Note that, this is exactly the same procedure as in~\cite{jabri2020walk}, 
we simply denote the propagation process as ``Mask-prop'' here:
\begin{equation}
    \label{mask-prop}
    \hat{M}_t = \text{Mask-prop}(\hat{M}_{t-1},\dots,\hat{M}_{t-n})
\end{equation}
where $\hat{M}_{t-1},\dots,\hat{M}_{t-n}$ refer to the  previously obtained results within a temporal window size $n$. 

In contrast to the conventional semi-supervised VOS scenario,
where the groundtruth mask at frame $0$ is given for propagation, in our work, 
\textbf{no} groundtruth segmentation mask is provided. Instead, we pick one key frame from our OCLR predictions, and then propagate the masks of those objects bi-directionally. The choice of the starting frame is based on the temporal coherence of the optical flow segmentation predictions. 
In the following, 
we describe the process for discovering frames with the most coherent predictions temporally:
\begin{enumerate}
    \item 
    We measure the temporal coherence by propagating our OCLR prediction~($\hat{M}^f_{t-1}$) to the next frame, 
    {\em i.e.} $\hat{M}_t =  \text{Mask-prop}(\hat{M}^f_{t-1})$,
    and compute the L1 distance between the propagation and our OCLR prediction at frame $t$, {\em i.e.}~$L_t = |\hat{M}^f_{t} - \hat{M}_t|$.
    \item We take a frame-wise average of L1 loss as $L_{\text{mean}} = \frac{1}{T} \sum_{t=1}^T L_t$, and find a set of temporally consistent frames $\{t_s\}$ with $L_{t_s} < L_{\text{mean}}$.
    \item Among the frame set $\{t_s\}$, one frame $t_k$ with the lowest L1 loss ({\em i.e.} $L_{t_k} = \min{(\{L_t\})}$) is then selected as the starting frame for mask propagations .
\end{enumerate}
The motivation behind this selection process is that, at frames with lower L1 loss, 
the predicted mask by our OCLR model is more consistent with those in adjacent frames. We then start a bi-directional mask propagation from the key frame $t_k$ with a temporal window size $n=7$.

\paragraph{Dynamic refinement.} 
Apart from using the prediction of frame $t_k$ as starting frame, we also fuse the other selected frames $\{t_s\}$ into the propagation process for refinement. Specifically, if a new frame $t$ belongs to the set $\{t_s\}$, the propagation result $\hat{M}_{t}$ would be averaged with the mask predicted by our flow-based model $\hat{M}^f_{t}$. The resultant mask, {\em i.e.} $(\hat{M}_{t}$ + $\hat{M}^f_{t})/2$, would then be applied to propagate later frames. 

By utilizing masks predicted by our flow-based model, the temporally accumulated drifting issue can be largely alleviated,
especially on the object boundaries, 
as they are usually clearly delineated in our OCLR flow-based model. We refer this process as dynamic refinement, short for \textbf{Dyn. Ref.}
\camera{\paragraph{Conditional random field (CRF).} Following the common practices in video object segmentation tasks, we adopt CRF as the final post-processing step that utilizes RGB information to refine mask predictions.}

\section{Datasets}

\subsection{Dataset details}
To evaluate our multi-layer model, 
we benchmark on multiple popular datasets on both single and multiple object segmentation tasks.

{\noindent \bf DAVIS2016~\cite{Perazzi16}} consists of $50$ high-resolution video sequences~($30$ for training and $20$ for validation) with $3455$ frames,
the primary moving objects in the scene have been annotated at the pixel level. We report our model performance on the validation set at a $480$p resolution.

{\noindent \bf SegTrackv2~\cite{FliICCV2013}} contains $14$ sequences and $947$ fully-annotated frames, with challenging cases such as occlusions, fast motion and complex shape deformations. 
Even though multiple objects may be annotated, 
the community often treats SegTrackv2 as a benchmark for single object segmentation~\cite{Jain17,yang_loquercio_2019}, by grouping objects in the foreground.

{\noindent \bf FBMS-59~\cite{OB14b}} contains $59$ sequences with a total of $720$ pixel-level annotations provided every $20$ frames. Similar to SegTrackv2, some multi-object sequences are relabelled for single object segmentation evaluations.

{\noindent \bf Moving Camouflaged Animals (MoCA)~\cite{Lamdouar20}} focuses on segmenting camouflaged animals moving in natural scenes. It contains $141$ high-resolution video sequences annotated by tight bounding boxes for every $5$th frame. Following~\cite{Yang21a}, we adopt a filtered MoCA dataset by excluding videos with predominantly no locomotion, resulting in $88$ video sequences and $4803$ frames

{\noindent \bf DAVIS2017~\cite{Ponttuset17}} extends DAVIS2016 dataset by introducing additional videos with multi-object contents, resulting in a total of 150 sequences with over $10$k pixel-level annotations. In particular, DAVIS2017 is a common benchmark for semi-supervised video object segmentation (VOS) tasks (i.e.\ with ground-truth first-frame annotation provided). In this work, the model performance is not directly evaluated on DAVIS2017, where there are sequences with common motion objects that are indistinguishable based on purely flow information. Instead, we adopt a curated DAVIS2017-motion dataset, with details provided in the next section.

Overall, we adopt a Hungarian matching process to associate layer predictions with the ground-truth annotations in DAVIS2016, MoCA, DAVIS2017-motion and our synthetic datasets. For evaluations on SegTrackv2 and FBMS-59, we instead group all layers together as a single foreground object.

\subsection{DAVIS2017-motion curation}
As objects in common motion cannot be distinguished purely from motion cues, we re-annotate the original DAVIS2017 dataset by grouping jointly moving objects to form a new DAVIS2017-motion dataset for benchmarking motion-based object segmentation. In this section, we first provide a definition of common motion, followed by some curation details.

\paragraph{Common motion.} 
In this work, the concept of object common motion is defined based on two necessary criteria: {\em~First}, objects are or appear to be spatially connected throughout the whole sequence. 
More specifically, pixel-level masks for different objects are next to each other for all frames. 
{\em~Second}, objects must share the same motion trends. 
A quick judgement can be made by observing if there is a noticeable flow discontinuity at the common boundary. 

\begin{figure}[!htb]
  \hspace{-0.7cm}
  \includegraphics[width=1.05\textwidth]{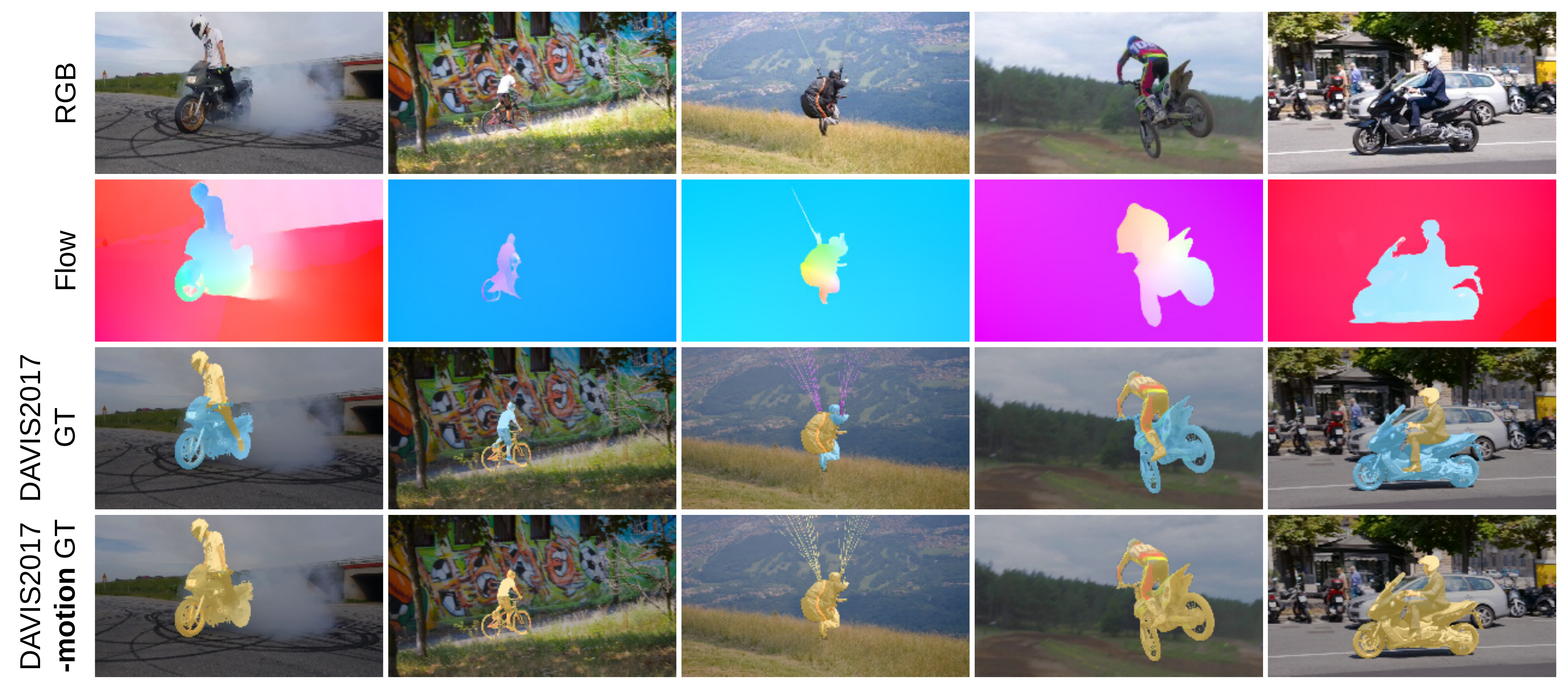}
  \caption[Curated DAVIS2017-motion dataset]{\textbf{Curated DAVIS2017-motion dataset.} 
  Note that, in the original DAVIS2017 dataset~(3rd row), 
  the objects have been annotated based on their semantic categories, however, in our curated DAVIS2017-motion dataset, we join those objects with common motion as a whole, for example, in the 1st, 4th, 5th columns, 
  the person and motorbike are originally annotated as two different classes, while in our curated dataset, they are labelled as a whole, as they follow the same motion.}
  \label{supfig:davis17m}
  \vspace{-0.5cm}
\end{figure}

\paragraph{Curation details.} 
As shown in Figure~\ref{supfig:davis17m}, 
based on the rules defined above, we group the annotations of jointly moving objects as a whole, resulting in the new DAVIS2017-motion dataset. 
Note that, as DAVIS2017-motion is adopted mainly for validation purposes, 
we re-annotate only the validation sequences in the original DAVIS2017. 
The full list of curated sequences includes: \textit{bmx-trees, horsejump-high, india, kite-surf, lab-coat, mbike-trick, motocross-jump, paragliding-launch,  scooter-black, shooting, soapbox}. 
We will release the curated dataset, 
together with the re-annotation codes for further research.

\section{Ablation study}
In this section, we present more details on our ablation studies,
with a single parameter varied every time.
Firstly, we conduct experiments on the pipeline for data simulation,
to get the best Sim2Real performance on single moving object segmentation;
Secondly, we extend the simulation procedure to multiple objects, 
and investigate more training details, for example, normalisation, loss function, number of frames, {\em etc}; 
Thirdly, we repeatedly train several motion segmentation models with the same optimal hyperparameters to verify our model stability.
\revised{Fourthly, we demonstrate the scalability of our method by training on an increasing number of synthetic frames. Finally, we show the effectiveness of our model by comparing it with other supervised models under the same supervision.}

\subsection{Synthetic dataset}

We simulate the videos with only a single object and conduct experiments to understand several key choices in the pipeline, for example, motion representation and background motion.
As shown in Table~\ref{sup:syn-table}, 
while comparing Ours-A and Ours-B, we observe that training on RAFT flows is beneficial, resulting in higher performance on all datasets. We conjecture that training on RAFT flows leads to a narrower Sim2Real domain gap, as flows in these test sets also come from RAFT. In addition, introducing real videos as backgrounds brings further performance gain, as shown in Ours-B and Ours-C. 
Therefore, we treat the settings in Ours-C as the default for our synthetic data pipeline.

\begin{table}[!htb]
\setlength\tabcolsep{8pt}
\footnotesize
  \centering
  \caption[Settings for synthetic data pipeline]{Settings for training in synthetic dataset pipeline.
  Note that, all the flows on test set are computed with RAFT.}
  \begin{tabular}{cccccc}  
    \toprule
    {} & \multicolumn{2}{c}{Synthetic Training Settings} & \multicolumn{3}{c}{$\mathcal{J}$ (Mean)  $\uparrow$}   \\
    \cmidrule(r){2-3}
    \cmidrule(r){4-6}
    Experiment    & Flow     & Background motion & DAVIS 2016  & SegTrackv2 & Syn-Single\\
    \midrule
    Ours-A & GT  & Homography  &  $67.4$ &  $52.3$ & $78.0$    \\
    Ours-B & RAFT & Homography & $68.6$&  $58.5$  & $84.7$    \\
    Ours-C & RAFT & Homography + Real video & $\textbf{72.0}$ &  $\textbf{62.3}$  & $\textbf{91.4}$    \\
    \bottomrule
  \end{tabular}
  \label{sup:syn-table}
\end{table}

\subsection{Training setting}

{\noindent \bf Instance normalisation. }  
As shown in Table~\ref{sup:trn-table}, 
while evaluating the multi-layer models ($N = 3$) on single object (DAVIS2016) and multi-object (DAVIS2017-motion, Syn-multi) segmentation datasets,
the model with instance normalizations~(Ours-G) consistently outperforms their counterparts~(Ours-D),
showing the importance of using instance normalization.

{\noindent \bf Hungarian matching. }
In our proposed architecture, 
the learnable object queries are permutation invariant, 
that is to say, each can correspond to different layers,
and we use Hungarian matching to assign the layer to each object query.
Here, we ablate the Hungarian matching procedure,
and instead force each query to predict a layer at fixed order, {\em e.g.}~1st query is always associated with the front layer.
As indicated by the result, 
such design significantly degrades the performance, 
as shown in Table~\ref{sup:trn-table} Ours-E.

{\noindent \bf Training by amodal segmentations. }
In Ours-F, 
we train the model only by modal masks in the synthetic dataset. 
The performance has dropped significantly, suggesting that explicit amodal supervision helps the network to learn object permanence within layers.

\begin{table}[!htb]
\setlength\tabcolsep{6pt}
\footnotesize
  \caption[Training parameters]{\textbf{Settings for training parameters.}~\textbf{IN}: Instance Normalisation;~\textbf{HM}: Hungarian matching; ~\textbf{Amodal}: Training on amodal mask (vs. modal mask); ~$\mathcal{\lambda}_{\textrm{bound}}$: weight on boundary loss; ~$T$: number of input frames. Syn-Val ($\mathcal{M}$|$\mathcal{A}$) corresponds to modal and amodal results on synthetic dataset. Ours-G denotes a default baseline setting compared to others.}
  \label{sup:trn-table}
  \centering
  \hspace{-.4cm}
  \begin{tabular}{ccccccccc}  
    \toprule
    {} & \multicolumn{5}{c}{Training settings} & \multicolumn{3}{c}{$\mathcal{J}$ (Mean)   $\uparrow$}   \\
    \cmidrule(r){2-6}
    \cmidrule(r){7-9}
    Model    & IN    & HM &  Amodal & $\mathcal{\lambda}_{\textrm{bound}}$ & $T$ & DAVIS2016 & DAVIS2017-motion & Syn-Val ($\mathcal{M}$|$\mathcal{A}$)\\
    \midrule
    Ours-D & \xmark  & $\checkmark$  &$\checkmark$  & $0.2$  & $30$ & $67.6$ &  $48.7$  &  $83.5$  | $83.0$   \\
    Ours-E & $\checkmark$ & \xmark   &$\checkmark$  & $0.2$  & $30$ &  $67.4$ &  $44.2$   & $70.6$ | $71.0$      \\
    Ours-F & $\checkmark$  & $\checkmark$  & \xmark  & $0.2$  & $30$ &  $69.2$ &  $50.5$  &  $81.1$ | $76.9$     \\ 
    Ours-G & $\checkmark$  & $\checkmark$  &$\checkmark$  & $0.2$  & $30$ & $\textbf{72.1}$ &  $\textbf{54.5}$  &  $\textbf{85.6}$ | $\textbf{84.7}$  \\ 
    \midrule
    Ours-H & $\checkmark$  & $\checkmark$  &$\checkmark$  & $0.2$  & $15$ & $71.3$ & $53.5$  &  $82.8$ | $83.0$  \\
    Ours-I & $\checkmark$  & $\checkmark$  &$\checkmark$  & $0$ & $30$ & $71.5$ & $54.1$ &  $80.9$ | $81.6$   \\ \bottomrule
  \end{tabular}
\end{table}
{\noindent \bf Boundary loss. }
While comparing between Ours-G and Ours-I,
we observe a performance boost from applying boundary loss. 
This validates our assumption that focusing on object boundaries can help the model to learn the information on object shapes and layer orders from optical flows.

{\noindent \bf Number of frames. }
Lastly, we compare our default model (Ours-G) with a variant that takes in a reduced number of input frames (Ours-H, $T=15$),
not surprisingly, a longer temporal input tends to also give slightly higher overall performance.

\revised{
\paragraph{Optical flow methods.} Since our OCLR model takes optical flows as the only input, the resultant performance is largely influenced by the quality of optical flow estimations, as demonstrated in Table~\ref{sup:opflow-table}. The highest performance of Our-G verifies our choice of the RAFT method for optical flow predictions.}

\begin{table}[!htb]
\setlength\tabcolsep{6pt}
\footnotesize
  \caption[Flow]{\revised{Choice of optical flow methods.}}
  \label{sup:opflow-table}
  \centering
  \hspace{-.4cm}
  {\color{black}\begin{tabular}{cccc}  
    \toprule
    {} & {} & \multicolumn{2}{c}{$\mathcal{J}$ (Mean)   $\uparrow$}   \\
    \cmidrule(r){3-4}
    Model    & Optical flow & DAVIS2016 & DAVIS2017-motion \\
    \midrule
    Ours-G & RAFT~\cite{Teed20}  &  $\textbf{72.1}$ &  $\textbf{54.5}$  
    \\
    Ours-J & ARFlow~\cite{liu2020learning} &  $54.6$ &  $39.5$  
    \\
    Ours-K & MaskFlownet~\cite{zhao2020maskflownet}  &  $66.0$ &  $49.0$  
    \\
    \bottomrule
  \end{tabular}}
\end{table}

\subsection{Repeating experiments}
In Table~\ref{sup:rep-table}, we demonstrate multi-object segmentation results by re-training the model several times based on the default setting (Ours-G in Table~\ref{sup:trn-table}). As can be observed, there are minimal performance differences between repeated experiments (within $1\%$), which validates the reliability of our results.

\begin{table}[!htb]
  \footnotesize
  \caption[Repeating experiments]{Repeating experiments on multi-object segmentation tasks.}
  \label{sup:rep-table}
  \centering
  \begin{tabular}{cccc}  
    \toprule
    {} & \multicolumn{3}{c}{$\mathcal{J}$ (Mean)  $\uparrow$}   \\
    \cmidrule(r){2-4}
    Experiment   & DAVIS2016 & DAVIS2017-motion & Syn-Val ($\mathcal{M}$|$\mathcal{A}$)\\ 
    \midrule
    1 & $72.1$ &  $54.5$  & $85.6$ | $84.7$  \\
    2 & $72.0$ &  $54.1$  & $85.2$ | $84.3$  \\
    3 & $72.8$ &  $54.3$  & $84.9$ | $84.5$  \\
    4 & $72.1$ &  $54.0$  & $85.9$ | $85.2$  \\
    \midrule
    Mean &  $72.3$ &  $54.2$  & $85.4$ | $84.7$  \\
    Std. &  $\pm 0.3$ &  $\pm 0.2$  & $\pm 0.4$ | $\pm 0.3$  \\
    \bottomrule
  \end{tabular}
\end{table}

\revised{
\subsection{Scalability of model performance}
Table~\ref{sup:scale-table} demonstrates how our model performance scales with the amount of synthetic data. As more synthetic frames are introduced during training, there is a clear increase in both DAVIS2016 and DAVIS2017-motion performance.
Moreover, Ours-O and Ours-P correspond to models that are directly trained on real datasets ({\em i.e.}~DAVIS2017-motion). Limited to the number of manual annotations, these models achieve lower performance than synthetic-supervised counterparts, particularly in multi-object segmentation. 
}

\begin{table}[!htb]
  \footnotesize
  \caption[Scalability]{\revised{\textbf{Model performance scales with the size of training sets.} ``with aug.'' stands for augmentations applied on the input flows including random cropping, rotations, jittering, dropouts, {\em etc.}}}
  \label{sup:scale-table}
  \centering
  {\color{black}\begin{tabular}{ccccc}  
    \toprule
    {} & {} &{}& \multicolumn{2}{c}{$\mathcal{J}$ (Mean)  $\uparrow$}   \\
    \cmidrule(r){4-5}
    Model & Training set & No. of training frames   & DAVIS2016 & DAVIS2017-motion \\ 
    \midrule
    Ours-L & Syn-train subset & $43.2$k & $69.6$ &  $51.2$   \\
    Ours-M & Syn-train subset & $69.1$k & $70.2$ &  $51.5$   \\
    Ours-N & Syn-train subset & $115.2$k & $71.2$ &  $53.7$   \\
    Ours-G & Syn-train & $138.2$k & $\textbf{72.1}$ &  $\textbf{54.5}$   \\
    \midrule
    Ours-O & DAVIS2017-motion & $4.2$k & $66.7$ &  $42.8$   \\
    Ours-P & DAVIS2017-motion & $4.2$k (with aug.) & $69.4$ &  $45.3$   \\
    \bottomrule
  \end{tabular}}
\end{table}

\revised{
\subsection{Effectiveness of the OCLR model}
Table~\ref{sup:eff-table} provides a comparison between our OCLR model and two other supervised models on multi-object segmentation. All methods take optical flow as the only input and are supervised by either the real dataset (DAVIS2017-motion) or our synthetic data (Syn-train). Motion Grouping (sup.) represents a supervised version of Motion Grouping~\cite{Yang21a}, while the standard Mask R-CNN~\cite{matterport_maskrcnn_2017} model adopted follows the default settings in the original paper, with a ResNet-50-FPN backbone trained from scratch.} \camera{In this case, the Mask R-CNN model takes optical flows as the only input, and is therefore referred to as Mask R-CNN (flow-only).}

\begin{table}[!htb]
  \footnotesize
  \caption[Effectiveness]{\revised{\textbf{Comparison of different models under real or synthetic supervision}. All models take optical flow at the only input. During inference, the 1st frame GT is not available ({\em~i.e.}, unsupervised VOS). In column Sup. (supervision), ``Syn.'' and ``Real'' represent synthetic data supervision and real-data supervision, respectively.}}
  \label{sup:eff-table}
  \centering
  {\color{black}\begin{tabular}{ccccc}  
    \toprule
    {} & {} & {} & {} & {$\mathcal{J}$ (Mean)  $\uparrow$} \\
    \cmidrule(r){5-5}
    Model & Sup. & Training set & No. of training frames  & DAVIS2017-motion \\ 
    \midrule
    Motion Grouping (sup.) & Real & Syn-train & $4.2$k &  $32.7$   \\
    Mask R-CNN (flow-only) & Real & Syn-train & $4.2$k &   $40.3$  \\
    OCLR (flow-only) & Real & Syn-train & $4.2$k &   $42.8$  \\
    \midrule
    Motion Grouping (sup.) & Syn. & DAVIS2017-motion & $138.2$k &   $44.9$ \\
    Mask R-CNN (flow-only) & Syn. & DAVIS2017-motion  & $138.2$k & $50.4$  \\
    OCLR (flow-only) & Syn. & DAVIS2017-motion & $138.2$k &   $\textbf{54.5}$   \\
    \bottomrule
  \end{tabular}}
\end{table}

\revised{
From Table~\ref{sup:eff-table}, it can be observed that: (i) Our OCLR outperforms both benchmark models under both supervision scenarios, particularly under synthetic supervision. When visualising the qualitative results, we found that Mask R-CNN demonstrates inferior performance in comparison to OCLR, particularly when there is noisy optical flow, temporally stationary objects, or heavy object deformations; while in contrast, OCLR is designed with the ability to infer amodal masks, and thus to handle situations with occlusion happening; (ii) Motion Grouping originally designed for self-supervision does not perform well when direct supervision is applied; (iii) Compared to supervision provided by a limited amount of real data ($4.2$k frames), scalable synthetic supervision ($138.2$k frames) leads to general performance improvements.
}

\subsection{Settings for test-time adaptation}
\label{sec:tta}

\paragraph{Mask propagation.} 
According to Table~\ref{sup:test-table}, results obtained by RGB-based mask propagation from a key frame (Ours-Q) surpass the performance of the flow-only model (Ours-G). 
This validates the benefits of introducing RGB information during test time.

\paragraph{Dynamic refinement.} In Ours-R, the dynamic refinement is applied to the propagation process by introducing segmentation outputs from our OCLR model. Consequently, this additional object mask information on average leads to a $2\%$ improvement in performance.

\paragraph{Finetuning of vision transformer} 
By comparing Ours-S with Ours-R, we notice a further performance boost by per-sequence finetuning of the DINO-pretrained vision transformer, and the settings in Ours-S contribute to the highest test-time adaptation result.

\paragraph{Conditional random field (CRF)}
Finally, by comparing between Ours-T with Ours-S, applying CRF as the post-processing step further improves the overall performance.

\begin{table}[!htb]
  \caption[Settings for test-time adaptation]{Settings for test-time adaptation.}
  \label{sup:test-table}
  \centering
  \begin{tabular}{cccccccccc}  
    \toprule
    {} & \multicolumn{4}{c}{Training Settings} & \multicolumn{2}{c}{$\mathcal{J}$ (Mean)  $\uparrow$}   \\
    \cmidrule(r){2-5}
    \cmidrule(r){6-7}
    Model    & Mask Prop.  & Dyn. Ref.& Fine-tuning & CRF & DAVIS2016 & DAVIS2017-motion \\
    \midrule
    Ours-G & \xmark & \xmark   & \xmark  &  \xmark & $72.1$ &  $54.5$    \\
    Ours-Q & $\checkmark$   & \xmark   & \xmark &\xmark &$76.5$ &  $60.9$     \\
    Ours-R & $\checkmark$  & $\checkmark$  & \xmark &  \xmark & $77.4$ &  $63.5$    \\
    Ours-S & $\checkmark$  & $\checkmark$   & $\checkmark$ & \xmark & $78.9$ &  $63.9$   \\
    \camera{Ours-T} & $\checkmark$  & $\checkmark$   & $\checkmark$ & $\checkmark$ & \camera{$\textbf{80.9}$} &  \camera{$\textbf{65.2}$}    \\
    \bottomrule
  \end{tabular}
\end{table}

\section{Quantitative results}
Table~\ref{suptab:single} summarizes performance on single object video segmentation benchmarks. For camouflaged object detection on MoCA, Table~\ref{suptab:moca} provides a more detailed quantitative comparison across different approaches. The multiple object video segmentation results on DAVIS2017-motion are shown in Table~\ref{suptab:multi}.

\begin{table}[!htbp]
\setlength\tabcolsep{6pt}
  \caption[Quantitative comparison on DAVIS2016, SegTrackv2 and FBMS-59]{Quantitative comparison single object video segmentation benchmarks. Results from more methods are quoted compared to the table in the main text. \revised{``HA'' stands for human annotations. In column Sup. (supervision), ``None'', ``Syn.'', ``Real'' represent self-supervision, synthetic data supervision, and real data supervision, respectively.} \textbf{\textit{Bold}} represents the state-of-the-art performance (excluding our test-time adaptation results, which are labelled as $\textcolor{blue}{blue}$ instead). For FBMS, results outside/inside brackets correspond to flow inputs with $\pm 1$/$\pm 3$ frame gaps.}
  \label{suptab:single}
  \centering
  \footnotesize
  \begin{tabular}{cccccccc}  
    \toprule
    {} & \multicolumn{4}{c}{Training Settings} & \multicolumn{3}{c}{$\mathcal{J}$ (Mean)  $\uparrow$ } \\
    \cmidrule(r){2-5}
    \cmidrule(r){6-8}
    Model    & \revised{HA} & \revised{Sup.}    & RGB &  Flow  & DAVIS2016 & SegTrackv2 & FBMS-59\\
    \midrule
    SAGE~\cite{sage} & \revised{\xmark}  & \revised{None} & $\checkmark$  &$\checkmark$  &  $42.6$ &  $57.6$  &  $61.2$     \\
    NLC~\cite{faktor2014videonlc} & \revised{\xmark} & \revised{None}  & $\checkmark$  &$\checkmark$  &  $55.1$ &  $67.2$  &  $51.5$  \\
    CUT~\cite{Keuper15} & \revised{\xmark}  & \revised{None} &  $\checkmark$  &$\checkmark$  &  $55.2$ &  $54.3$  &  $57.2$     \\
    FTS~\cite{Papazoglou13} & \revised{\xmark}  & \revised{None} & $\checkmark$  &$\checkmark$  &  $55.8$ &  $47.8$  &  $47.7$     \\
    CIS~\cite{yang_loquercio_2019}& \revised{\xmark} & \revised{None} & $\checkmark$  &$\checkmark$  &  $59.2$ &  $45.6$  &  $36.8$ \\
    CIS (w. post-process.)~\cite{yang_loquercio_2019} & \revised{\xmark} & \revised{None} & $\checkmark$  &$\checkmark$  &  $71.5$ &  $62.0$  &  $63.5$   \\
    Motion Grouping~\cite{Yang21a} & \revised{\xmark} & \revised{None} & \xmark  &$\checkmark$  &  $68.3$ &  $58.6$  &  $53.1$ \\
    SIMO~\cite{Lamdouar21} & \revised{\xmark} & \revised{Syn.} & \xmark  &$\checkmark$  &  $67.8$ &  $62.0$  &  $-$     \\
    \textbf{OCLR (flow-only)}& \revised{\xmark} & \revised{Syn.}  & \xmark  &$\checkmark$  &  $\textbf{72.1}$ &  $\textbf{67.6}$  &  $\textbf{65.4 (70.0)}$ \\
    \camera{\textbf{OCLR (test. adap.)}} & \revised{\xmark} & \revised{Syn.}  &$\checkmark$  &$\checkmark$  &  \camera{$\textcolor{blue}{80.9}$} &  \camera{$\textcolor{blue}{72.3}$}  &  $\textcolor{blue}{69.8}$ \textcolor{blue}{(}$\textcolor{blue}{72.7}$\textcolor{blue}{)} \\
    \midrule
    SFL~\cite{Cheng17} & \revised{$\checkmark$} & \revised{Real} & $\checkmark$  & $\checkmark$  &  $67.4$ &  $-$  &  $-$ \\
    FSEG~\cite{Jain17} & \revised{$\checkmark$}  & \revised{Real} & $\checkmark$  & $\checkmark$  &  $70.7$ &  $\textbf{61.4}$  &  $68.4$ \\
    LVO~\cite{Tokmakov19} & \revised{$\checkmark$}  & \revised{Real} & $\checkmark$  & $\checkmark$  &  $75.9$ &  $57.3$  &  $65.1$ \\
    ARP~\cite{song2018pyramidpdb} & \revised{$\checkmark$}  & \revised{Real} & $\checkmark$  & $\checkmark$  &  $76.2$ &  $57.2$  &  $59.8$ \\
    COSNet~\cite{Lu_2019_CVPR} & \revised{$\checkmark$}  & \revised{Real} & $\checkmark$  & \xmark  &  $80.5$ &  $49.7$   &  $75.6$    \\
    MATNet~\cite{zhou20} & \revised{$\checkmark$}  & \revised{Real} & $\checkmark$  &$\checkmark$  &  $82.4$  &  $50.4$   &  $\textbf{76.1}$    \\
    3DC-Seg~\cite{Mahadevan20BMVC3dc} & \revised{$\checkmark$}  & \revised{Real} & $\checkmark$  &$\checkmark$  &  $84.3$  &  $-$   &  $-$   \\
    D$^2$Conv3D~\cite{D2conv3d} & \revised{$\checkmark$}  &  \revised{Real} & $\checkmark$  & \xmark  &  $\textbf{85.5}$  &  $-$   &  $-$   \\
    \bottomrule
  \end{tabular}
\end{table}

\begin{table}
\footnotesize
\setlength\tabcolsep{6pt}
  \caption[Quantitative comparison on MoCA]{Quantitative comparison of camouflaged object detection on MoCA. \revised{``HA'' stands for human annotations. In column Sup. (supervision), ``None'', ``Syn.'', ``Real'' represent self-supervision, synthetic data supervision, and real data supervision, respectively.} \textbf{\textit{Bold}} represents the state-of-the-art performance.}
  \label{suptab:moca}
  \hspace{-1.3cm}
  \begin{tabular}{cccccccccccc}  
    \toprule
    {} & \multicolumn{4}{c}{Training settings} & {} & \multicolumn{6}{c}{Detection Success Rate  $\uparrow$}   \\
    \cmidrule(r){2-5}
    \cmidrule(r){7-12}
    Model  & \revised{HA}  & \revised{Sup.}    & RGB &  Flow  & {$\mathcal{J}$  $\uparrow$} & $\tau = 0.5$ & $\tau = 0.6$& $\tau = 0.7$& $\tau = 0.8$& $\tau = 0.9$& mean\\
    \midrule
    CIS~\cite{yang_loquercio_2019} & \revised{\xmark}  & \revised{None} & $\checkmark$  &$\checkmark$  &  $49.4$ &  $0.556$  &  $0.463$ &  $0.329$ &  $0.176$ &  $0.030$ &  $0.311$    \\
    CIS (w. post-process.) & \revised{\xmark} & \revised{None} & $\checkmark$  &$\checkmark$  &  $54.1$ &  $0.631$  &  $0.542$ &  $0.399$  &  $0.210$ &  $0.033$ &  $0.363$   \\
    Motion Grouping~\cite{Yang21a} & \revised{\xmark}  & \revised{None} & \xmark  &$\checkmark$  &  $63.4$ &  $0.742$  &  $0.654$ &  $0.524$  &  $0.351$ &  $0.147$ &  $0.484$     \\
    SIMO~\cite{Lamdouar21} & \revised{\xmark} & \revised{Syn.}  & \xmark  &$\checkmark$  &   $68.6$ &  $0.772$  &  $0.717$ &  $0.623$  &  $0.464$ &  $0.255$ &  $0.566$  \\
    \textbf{Ours (flow-only)} & \revised{\xmark} & \revised{Syn.} & \xmark  &$\checkmark$  &   $\textbf{70.9}$   &  $\textbf{0.795}$  & $\textbf{0.743}$ &  $\textbf{0.658}$  &  $\textbf{0.508}$ & $\textbf{0.289}$  & $\textbf{0.599}$   \\
    \camera{\textbf{Ours (test. adap.)}} & \revised{\xmark} & \revised{Syn.} &$\checkmark$  &$\checkmark$  &   \camera{$67.5$}   &  \camera{$0.789$}  &  \camera{$0.717$} &  \camera{$0.615$}  &  \camera{$0.445$} &  \camera{$0.230$} & \camera{$0.559$}  \\
    \midrule
    COD~\cite{Lamdouar20} & \revised{$\checkmark$}  & \revised{Real} & \xmark  & $\checkmark$  & $44.9$ &  $0.414$  &  $0.330$ &  $0.235$  &  $0.140$ &  $0.059$ &  $0.236$    \\
    COD (two-stream) & \revised{$\checkmark$} & \revised{Real} & $\checkmark$  & $\checkmark$  & $55.3$ &  $0.602$  &  $0.523$ &  $0.413$  &  $0.267$ &  $0.088$ &  $0.379$  
      \\
    COSNet~\cite{Lu_2019_CVPR} & \revised{$\checkmark$} & \revised{Real} & $\checkmark$  & \xmark  & $50.7$ &  $0.588$  &  $0.534$ &  $0.457$  &  $0.337$ &  $0.167$ &  $0.417$       \\
    MATNet~\cite{zhou20} & \revised{$\checkmark$} & \revised{Real} & $\checkmark$  &$\checkmark$  & 64.2  & 0.712 &  0.670 & 0.599  &  0.492 &  0.246 &  0.544 \\
    \bottomrule
  \end{tabular}
\end{table}

\begin{table}[!htb]
\footnotesize
\setlength\tabcolsep{9pt}
  \caption[Quantitative comparison on DAVIS2017-motion.]{Quantitative comparison of multi-object video segmentation on DAVIS2017-motion. Results for semi-supervised methods are obtained by re-running source codes on the DAVIS2017-motion. \camera{Note that, the compared methods here are trained without using any human annotations during training, in particular, Motion Grouping (sup.), Mask R-CNN (flow-only) and OCLR models are supervised by only synthetic data, and other approaches are trained with self-supervision.} \textbf{\textit{Bold}} represents the state-of-the-art performance (excluding our test-time adaptation results, which are labelled as $\textcolor{blue}{blue}$ instead).}
  \label{suptab:multi}
  \hspace{-.35cm}
  \begin{tabular}{ccccccc}  
    \toprule
    {} & \multicolumn{3}{c}{Training settings} & \multicolumn{3}{c}{DAVIS2017-motion performance } \\
    \cmidrule(r){2-4}
    \cmidrule(r){5-7}
    Model    & {1st-frame-GT} & RGB &  Flow  & {$\mathcal{J}$\&$\mathcal{F}$  $\uparrow$} & {$\mathcal{J}$ (Mean)  $\uparrow$} & {$\mathcal{F}$ (Mean)  $\uparrow$}\\
    \midrule
    Motion Grouping~\cite{Yang21a} & \xmark & \xmark  &$\checkmark$  &  $35.8$ & $38.4$  &  $33.2$    \\
    \revised{Motion Grouping (sup.)} & \revised{\xmark} & \revised{\xmark}  & \revised{$\checkmark$}  &  \revised{$39.5$} & \revised{$44.9$}  &  \revised{$34.2$}    \\
    \camera{Mask R-CNN (flow-only)} & \camera{\xmark} & \camera{\xmark}  & \camera{$\checkmark$}  &  \camera{$50.3$} & \camera{$50.4$}  &  \camera{$50.2$}    \\
    \textbf{OCLR (flow-only)} &  \xmark  &  \xmark  &$\checkmark$  &  $\textbf{55.1}$ &  $\textbf{54.5}$  &  $\textbf{55.7}$    \\
    \camera{\textbf{OCLR (test. adap.)}} & \xmark  & $\checkmark$  & $\checkmark$  &  $\textcolor{blue}{64.4}$ &  $\textcolor{blue}{65.2}$  &  $\textcolor{blue}{63.6}$  \\
    \midrule
    CorrFlow~\cite{Lai19} & $\checkmark$  & $\checkmark$  &  \xmark &  $54.0$ &  $54.2$  &  $53.7$     \\
    UVC~\cite{nips19_joint_task} & $\checkmark$  & $\checkmark$  &  \xmark &  $65.5$ &  $66.2$  &  $64.7$ \\
    MAST~\cite{Lai20} & $\checkmark$  & $\checkmark$  &  \xmark &  $70.9$ &  $71.0$  &  $70.8$    \\
    CRW~\cite{jabri2020walk} & $\checkmark$  & $\checkmark$  & \xmark &  $73.4$ &  $72.9$  &  $74.1$     \\
    MAMP~\cite{Miao2022mamp} & $\checkmark$  & $\checkmark$  &$\checkmark$  &  $75.8$ &  $76.4$  &  $75.2$     \\
    DINO~\cite{caron2021emerging} & $\checkmark$  & $\checkmark$  &  \xmark  & $\textbf{78.7}$ &  $\textbf{77.7}$  &  $\textbf{79.6}$  \\
    \bottomrule
  \end{tabular}
\end{table}

\section{Qualitative results}
\begin{figure}
  \hspace{-1.0cm}
  \includegraphics[width=1.1\textwidth]{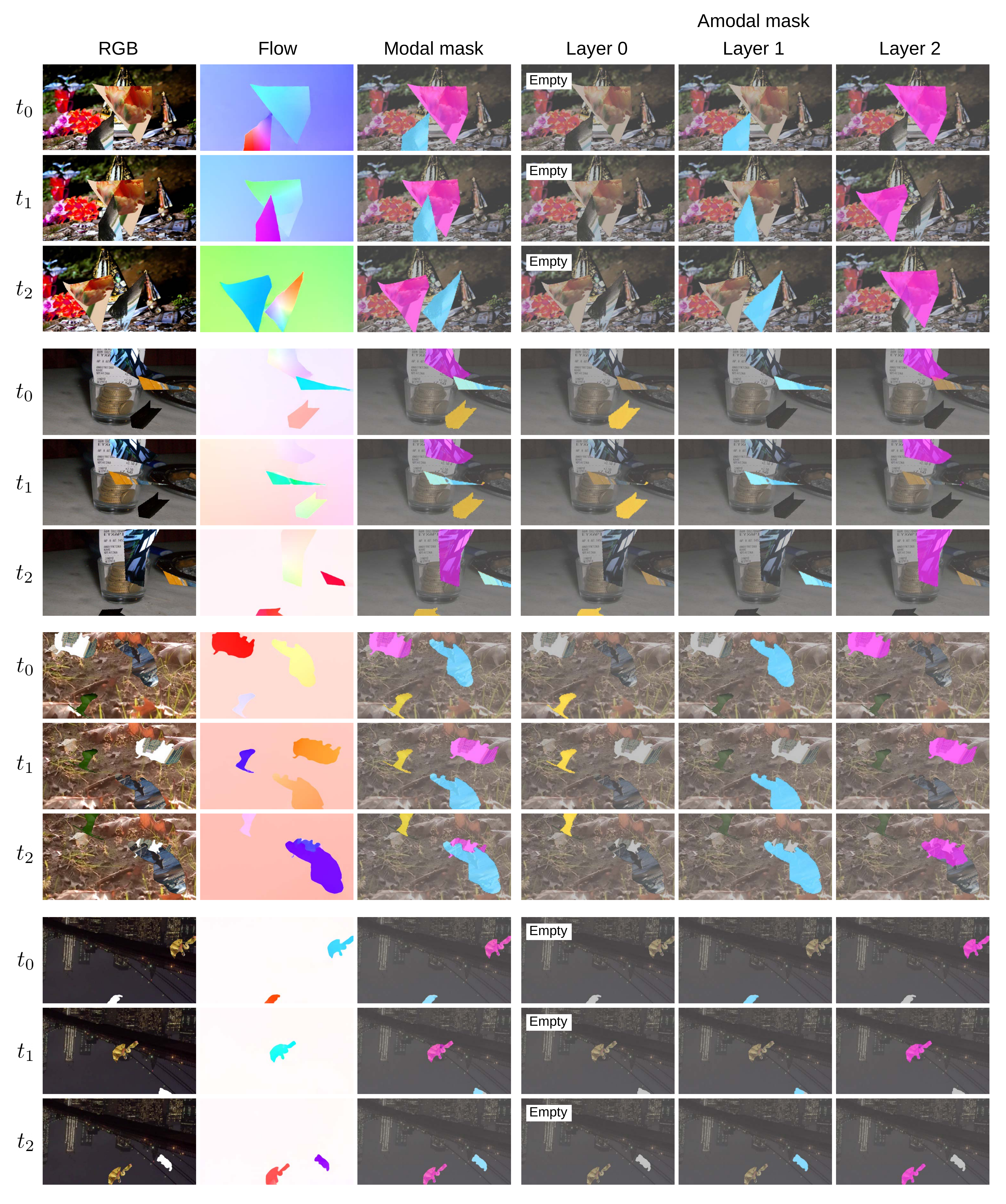}
  \caption[Qualitative results on our synthetic dataset Syn-Val]{Qualitative results on our synthetic dataset Syn-Val. Both modal and layer-wise amodal predictions are shown.}
  \label{supfig:syn_result}
\end{figure}

\begin{figure}
  \hspace{-0.7cm}
  \includegraphics[width=1.05\textwidth]{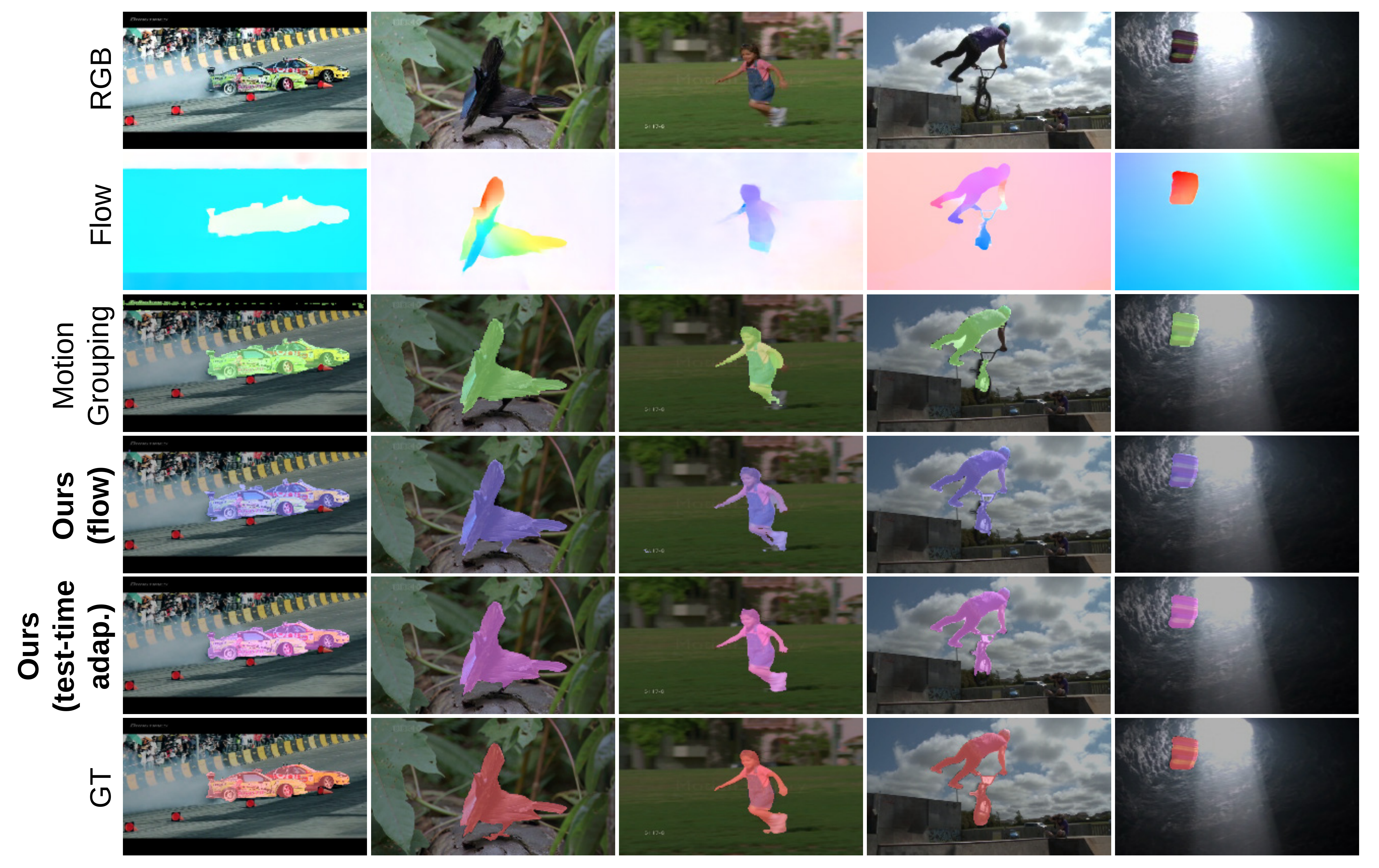}
  \caption[Qualitative results on SegTrackv2]{Qualitative results of single object video segmentation on SegTrackv2.}
  \label{supfig:segtrack}
\end{figure}

\begin{figure}
  \hspace{-0.7cm}
  \includegraphics[width=1.05\textwidth]{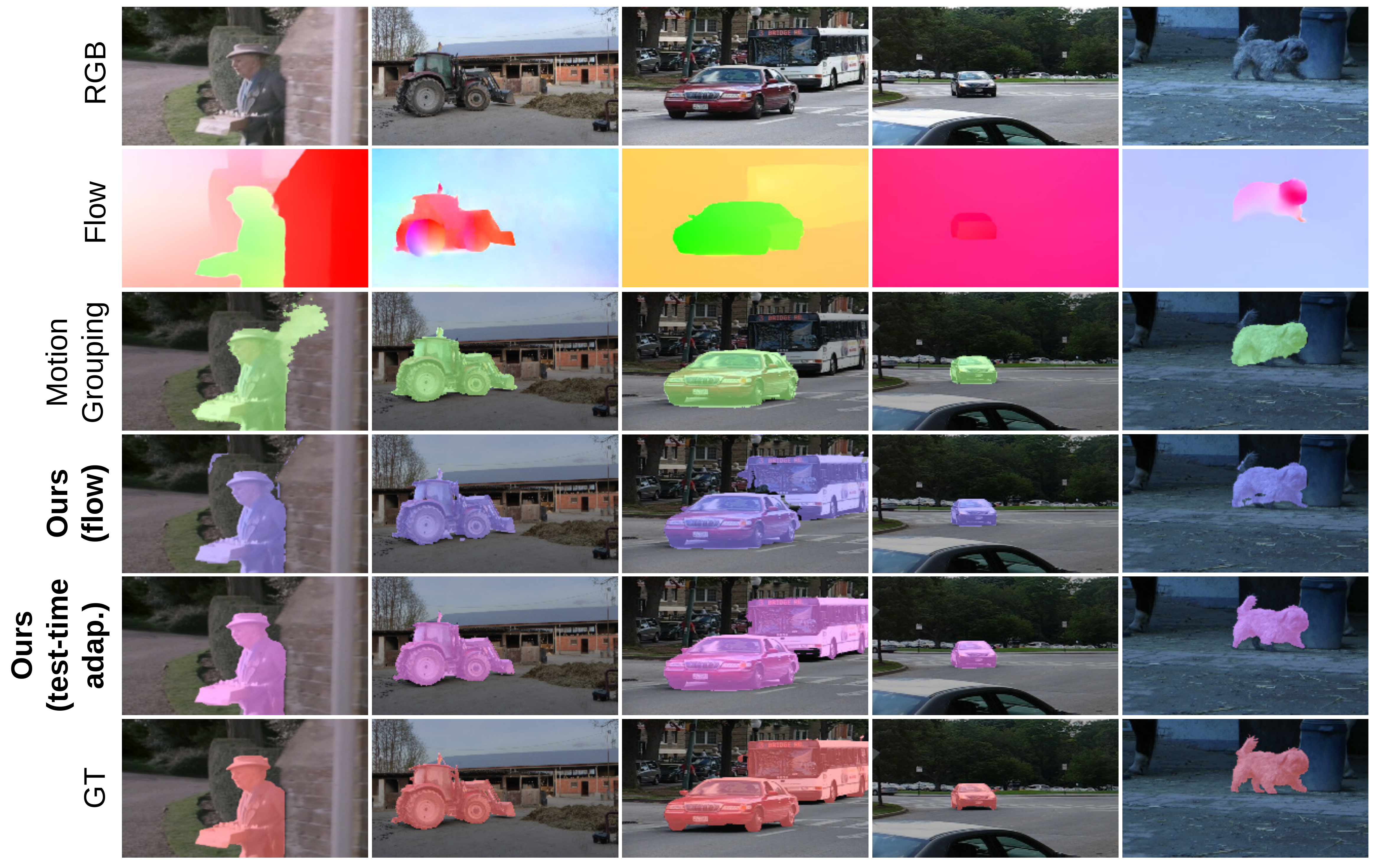}
  \caption[Qualitative results on FBMS]{Qualitative results of single object video segmentation on FBMS.}
  \label{supfig:fbms}
\end{figure}

Figure~\ref{supfig:syn_result},~\ref{supfig:segtrack} and~\ref{supfig:fbms} illustrate our model predictions on Syn-Val, SegTrackv2 and FBMS-59, respectively.
We also demonstrates qualitative results on synthetic and real datasets in the supplementary video.

\section{Discussions on ethic guideline}
\label{appendix:ethic}
\subsection{Potential negative societal impacts}

In our work, the main source of information is from our synthetic dataset, which is generated based on textures from the PASS dataset (without any personally identifiable information) and only shapes from the YouTubeVOS2018 training sets. These procedures ensure that almost no human information is introduced to our network. Furthermore, we advocate a segmentation method mainly utilizing motion cues, represented by optical flows. The textures in optical flow are only related to the motion fields and object shapes, without any real-world semantic information. By filtering out possible human-related information, we have made largely eliminated any social negative impacts on the environment, human rights, economics, personal security,{\em~etc.}

\subsection{General ethical conduct guideline}

\paragraph{Does the work contain any personally identifiable information or sensitive personally identifiable information?} 
No. The dataset we benchmark and curate,{\em~e.g.} DAVIS, SegTrackv2, FBMS and MoCA, do not contain sensitive personally identifiable information. For the generation of our synthetic dataset, we obtain all RGB texture information from the PASS dataset, which does not contain any personally identifiable information. For shape information, we use randomly generated polygon or only real-object shapes from the YouTubeVOS2018 training sets, without any textural details, which greatly eliminates possible sensitive information. If there is any personally identifiable information later found in any of the above datasets, we will make immediate corresponding changes. 

\paragraph{Does the work contain information that could be deduced about individuals that they have not consented to share?} 
No. As described above, all information provided to train and test our model does not contain sensitive personally identifiable information. Therefore, no personal privacy information could be deduced.

\paragraph{Does the work encode, contain, or potentially exacerbate bias against people of a certain gender, race, sexuality, or who have other protected characteristics?} 
No. As explained above, we tried to eliminate personally identifiable information in our synthetic dataset, which in turn minimises possible human-related biases.

\paragraph{Does the work contain human subject experimentation and whether it has been reviewed and approved by a relevant oversight board?}
No. We did not include human subject experimentation in our work.

\paragraph{Have the work been discredited by the creators?} 
No. All datasets we used and curated are under the CC-BY license, and we make necessary references to the original source.

\paragraph{Consent to use or share the data. Explain whether you have asked the data owner’s permission to use or share data and what the outcome was.}
As explained above, all adopted datasets follow the CC-BY license, and we have made the necessary references. 

\paragraph{Do you have domain specific considerations when working with high-risk groups.}
No. We minimise personally identifiable information in our synthetic dataset, therefore not raising any domain-specific issues regarding high-risk groups.

\paragraph{Have you filtered offensive content.} 
Yes. As mentioned above, we filter out offensive content in our synthetic dataset.

\paragraph{Can you guarantee compliance to GDPR?} 
Yes.

\end{document}